\documentclass[journal]{IEEEtran}
\usepackage{color}
\newcommand{\red}[1]{\textcolor[rgb]{0.5,0,0}{#1}}           % Color roig
\usepackage{amsfonts}
\usepackage{amssymb}
\usepackage{amsbsy} % Per a \boldsymbol
\usepackage{amsmath}
\usepackage{graphicx}
\usepackage{epstopdf}
\usepackage{amsthm}
\usepackage{bbm}

\usepackage{flushend} %balance columns in last page

\usepackage{changes}
\definechangesauthor[name={Daniel}, color=orange]{dan}

\usepackage{array}
\newcolumntype{L}[1]{>{\raggedright\let\newline\\\arraybackslash\hspace{0pt}}m{#1}}
\newcolumntype{C}[1]{>{\centering\let\newline\\\arraybackslash\hspace{0pt}}m{#1}}
\newcolumntype{R}[1]{>{\raggedleft\let\newline\\\arraybackslash\hspace{0pt}}m{#1}}

\usepackage{pgffor} % for 
\usepackage{float} % avoid floats error latex
\usepackage{wrapfig}
\usepackage{comment}
\usepackage{algorithm}
\usepackage[noend]{algpseudocode}
\usepackage{amsmath}
\usepackage{amssymb}
\usepackage{amsthm}
\usepackage{mathtools}
\usepackage{array}
\usepackage{makecell}
\usepackage{placeins}
\usepackage{multirow}
\usepackage{url}
\usepackage{cite}
\usepackage{wrapfig,lipsum,booktabs}
\usepackage[font=small,labelfont=bf]{caption}
\usepackage{graphicx}
\usepackage{colortbl}
\usepackage{xcolor}
\usepackage[all]{xy}
\usepackage{adjustbox}
\usepackage{hyperref}
\hypersetup{
    unicode=false,          % non-Latin characters in Acrobat’s bookmarks
    pdftoolbar=true,        % show Acrobat’s toolbar?
    pdfmenubar=true,        % show Acrobat’s menu?
    pdffitwindow=false,     % window fit to page when opened
    pdfstartview={FitH},    % fits the width of the page to the window
    pdfauthor={Camps-Valls},     % author
    pdfcreator={Camps-Valls},   % creator of the document
    pdfproducer={Camps-Valls}, % producer of the document
    pdfkeywords={}, % list of keywords
    pdfnewwindow=true,      % links in new window
    colorlinks=true,        % false: boxed links; true: colored links
    linkcolor={blue},          % color of internal links (change box color with linkbordercolor)
    citecolor=blue,         % color of links to bibliography
    filecolor=blue,         % color of file links
    urlcolor=blue           % color of external links
}

\def\Real{{\mathbbm{R}}}

\usepackage{pgf}
\usepackage{tikz}
\usetikzlibrary{arrows,automata,positioning,shapes}

\tikzset{
    >=stealth',
    punkt/.style={
           rectangle,
           rounded corners,
           draw=black, very thick,
           text width=6.5em,
           minimum height=2em,
           text centered},
    pil/.style={
           ->,
           thick,
           shorten <=2pt,
           shorten >=2pt,}
}

\newcommand{\Normal}[0]{\mathcal{N}}

\newcommand{\I}{\mathbf{I}}

\newcommand{\bK}{\mathbf{K}}

\newcommand{\bk}{\mathbf{k}}

\newcommand{\y}{\mathbf{y}}

\newcommand{\mse}{\textrm{MSE}}
\newcommand{\rmse}{\textrm{NMSE}}
\newcommand{\lai}{\textrm{LAI}}
\newcommand{\fapar}{\textrm{FAPAR}}
\newcommand{\thetavec}{\boldsymbol{\theta}}

\newcommand{\diff}{\textnormal{d}}

\renewcommand{\vec}[1]{\mathbf{#1}}

\title{Integrating Domain Knowledge in Data-driven Earth Observation with Process Convolutions}%: Latent Force Models for Biophysical Parameter Retrieval} % Is it even retrieval when we are not mapping from spectrum to biophysical parameter?

\usepackage{fancyhdr}

\fancypagestyle{firststyle}
{
   \fancyhf{}
   \fancyhead[C]{ \textsuperscript{\textcopyright}IEEE. Accepted for publication IEEE TGRS. DOI 10.1109/TGRS.2021.3059550  }
}

\author{Daniel Heestermans Svendsen,
Maria Piles,~\IEEEmembership{Senior Member~IEEE},
Jordi Mu{\~n}oz-Mar\'i,~\IEEEmembership{Member~IEEE}, \\
David Luengo,~\IEEEmembership{Senior Member~IEEE},
Luca Martino and Gustau Camps-Valls,~\IEEEmembership{Fellow~IEEE}
\thanks{Research funded by the European Research Council (ERC) under the ERC-CoG-2014 SEDAL (grant agreement 647423) and the USMILE-SyG-2019 (grant agreement 855187), and by projects KERMES (TEC2016-81900-REDT) and LEAVES (RTI2018-096765-A-100,MCIU/AEI/FEDER, UE). M. Piles is supported by a Ram\'on y Cajal contract (MICINN).}
\thanks{DHS, MP, JMM and GCV are with the Image Processing Laboratory, Universitat de Val\'{e}ncia, Spain.  \url{http://isp.uv.es/}, e-mail: \url{daniel.svendsen@uv.es}. DL is with the Universidad Polit\'ecnica de Madrid, Spain. e-mail: \url{david.luengo@upm.es}. LM is with Universidad Rey Juan Carlos, Spain. e-mail: \url{luca.martino@urjc.es}}
}

\begin{document}

\maketitle

\thispagestyle{firststyle}

\begin{abstract}
The modelling of Earth observation data is a challenging problem, typically approached by either purely mechanistic or purely data-driven methods. Mechanistic models encode the domain knowledge and physical rules governing the system. Such models, however, need the correct specification of all interactions between variables in the problem and the appropriate parameterization is a challenge in itself. On the other hand, machine learning approaches are flexible data-driven tools, able to approximate arbitrarily complex functions, but lack interpretability and struggle when data is scarce or in extrapolation regimes. 
In this paper, we argue that {\em hybrid learning schemes} that combine both approaches can address all these issues efficiently. We introduce {\em Gaussian process (GP) convolution models} for hybrid modelling in Earth observation (EO) problems. We specifically propose the use of a class of GP convolution models called {\em latent force models} (LFMs) for EO time series modelling, analysis and understanding. LFMs are hybrid models that incorporate physical knowledge encoded in differential equations into a multioutput GP model. LFMs can transfer information across time-series, cope with missing observations, infer explicit latent functions forcing the system, and learn parameterizations which are very helpful for system analysis and interpretability. 
We illustrate the performance in two case studies. First, we consider time series of soil moisture from active (ASCAT) and passive (SMOS, AMSR2) microwave satellites. We show how assuming a first order differential equation as governing equation, the model automatically estimates the e-folding time or decay rate related to soil moisture persistence and discovers latent forces related to precipitation. 
In the second case study, we show how the model can fill in gaps of leaf area index (LAI) and Fraction of Absorbed Photosynthetically Active Radiation (fAPAR) from MODIS optical time series by exploiting 
their relations across different spatial and temporal domains. The proposed hybrid methodology reconciles the two main approaches in remote sensing parameter estimation by blending statistical learning and  mechanistic modeling.
\end{abstract}


\vspace{1cm}
\section{Introduction} \label{sec:intro}

% ------------------------------------------------------
\iffalse
    \red{\scriptsize GUS, 2 pages. The outline could be:
    \begin{itemize}
    \item EO poses challenging problems of time series analysis: gaps, heterosc noise, structured data, multitoutput
    \item ML and SP plays a significant role in tackling these problems
    \item Many regression models deployed, yet recently GPs showed great success
    \item However ... Models should be nonlinear, deal with gaps, and respect the physics!
        \item Encode rules (via ODEs)
        \item Learn something about the model!
        \item Cope with multiple sensors and multiple models simultaneously!
        \item We introduce a little monster to do it altogether...
    \item GPs offer advantages: solid Bayesian framework, and covariance design to cope with specificities of data
    \item We will show examples in several problems involving land apps (SM, LAI, fAPAR,) and with different sensors (optical, microwave) 
    \item Outline
    \end{itemize}
    }
\fi
% ------------------------------------------------------

% Process understanding and modeling is at the core of the scientific reasoning. Principled {\em parametric} and mechanistic modeling has dominated science and engineering until recently with the advent of machine learning. Despite great success in many areas, machine learning algorithms in Earth observation faces the problem of credibility and physical consistency. %, as often do not respect the most elementary laws of physics. 

\IEEEPARstart{E}{arth} observation data analysis requires dealing with heterogeneous data and models, both statistical and process-based ones. Physicists and environmental scientists attempt to model systems in a principled way through analytic descriptions that encode prior beliefs of the underlying processes. Conservation laws, physical principles or phenomenological behaviors are generally formalized using mechanistic models and differential equations. This physical paradigm has been, and still is, the main framework for modeling complex natural phenomena like e.g. those involved in Earth sciences, both in geosciences and in remote sensing. With the availability of very large datasets captured with different sensory systems, the physical modeling paradigm has been challenged (and in many cases replaced) by the statistical machine learning (ML) paradigm, which offers excellent accuracy and an amenable and prior-agnostic approach \cite{Reichstein19nat,Halevy09,Lipton18,Bezenac18iclr}. 

% - ML models however do not respect physics in general
Machine learning models can fit observations very well, but predictions may be inconsistent with the laws of physics or even physically meaningless (e.g. negative density estimations). This has been perhaps the most important criticism to ML algorithms, and a relevant reason why, historically, physical modelling and machine learning have often been treated as entirely separate fields under very different scientific paradigms (theory-driven versus data-driven). 
Likewise, there is an on-going debate about the limitations of traditional methodological frameworks: about their rigidity and lack of accuracy in general \cite{Halevy09,Lipton18}, and in the fields of geosciences and hydrology in particular\cite{Karpatne17}. 
% - Hybrid models are an alternative: many options, often too complex and heuristic
Recently, however, integration of domain knowledge and enforcing of physical consistency by encoding governing physical equations in ML models have been proposed as a principled way to provide strong theoretical constraints on top of the observational ones\cite{Reichstein19nat}. 
The synergy between the two approaches has been gaining attention, by either redesigning the model's architecture, augmenting the training dataset with simulations, or by including physical principles as constraints in the cost function to be optimized\cite{Karpatne17,jia2019physics,muralidhar2018incorporating,Reichstein19nat}. %We will focus on the latter approach.
In recent years, applications have been introduced in various subfields of the Earth sciences: in climate science \cite{faghmous2014big,o2018using}, turbulence modeling \cite{mohan2018deep,bode2019using,xiao2019reduced}, hydrology \cite{xu2015data} and remote sensing parameter retrieval\cite{svendsen17jgp}.

Among the many ML models available, Gaussian processes (GPs) \cite{Rasmussen06} are one of the most popular probabilistic models \cite{Ghahramani15nature}. 
GPs provide a flexible and nonparametric way to model the functional relationships between a set of potentially explanatory covariates and observations. GPs have excelled in Earth science problems in recent years, mainly for model inversion, forward modeling and emulation of complex codes, as well as for causal inference from observational data \cite{CampsValls16grsm,CampsValls19nsr,svendsen2020active}. 
GPs cast the regression problem in a probabilistic framework with solid mathematical properties, thus providing not only accurate estimates but also principled uncertainty estimates for the predictions, allowing for a formal treatment of uncertainty quantification. Besides, GPs readily accommodate data stemming from different sources (e.g. multiple sensors, platforms and simulations) \cite{Vasudevan12,CampsValls16grsm,CampsValls18sciasi}, allow the ranking of the covariates and gaining insight on the underlying process by using appropriate kernel functions \cite{Rasmussen06,duvenaud13,Cheng19natcom}, and due to their solid Bayesian formalism, GPs can also be designed to incorporate prior knowledge easily  \cite{Rasmussen06,alvarez2013linear,CampsValls16grsm}.

The main motivation of this work is to reconcile data-driven models with mechanistic modeling by introducing a nonparametric GP model for biophysical parameter estimation in remote sensing applications. 
Despite the success of GPs for remote sensing data modeling and analysis, various limitations can be identified. 
First, GP models have mostly been applied in a static fashion~\cite{gewali2019gaussian}, without encoding the relevant temporal information conveyed by phenological cycles of vegetation, or by the differential equations governing the carbon or water uptakes, for instance. Second, the few time-based covariances available nowadays are mere parametrizations encoding simple {\em ad hoc} features such as trends and periodicities~\cite{mackay1998introduction}, hence no richer relations are learned from the data. Third, few attempts of multioutput and structured GP models are available, which hampers adoption of these models in real EO problems. Finally, we should note that, very often, the EO time series show uneven sampling because of acquisition artifacts and the presence of clouds~\cite{mahajan2019cloud}. The latter is typically solved by parametric gap filling interpolation, which alters the time series structure by over-smoothing the series, thus producing unrealistic results. We want to address these problems by drawing predictive power from various related time series (e.g. from different sensors or from different sites) while being physically consistent with an underlying ordinary differential equation (ODE) generating process.

In this paper, we use \textit{Gaussian process convolution models} \cite{ver1998constructing} in order to address the above-mentioned problems. We show that {\em latent force model} (LFM), a type of process convolution, can address these problems simultaneously. LFMs, originally introduced in~\cite{alvarez2009latent,alvarez2013linear}, perform a convolution between a underlying (latent) Gaussian process functions and a smoothing kernel derived from the ODE assumed to govern the system. Hence, The LFM elegantly combines a data-driven modelling approach with a physics-driven model of the underlying system. The LFM presented here performs multioutput regression, adapts to the signal characteristics, is able to cope with missing data in the time series, and provides explicit latent functions that allow system analysis and evaluation.

We provide empirical evidence of performance in two challenging case studies involving bio-geophysical land parameters --soil moisture (SM), leaf area index (LAI), fraction of absorbed photosynthetic active radiation (fAPAR)-- estimated from different (microwave and optical) satellite sensors. 
First, we consider the estimation of soil moisture from active (ASCAT) and passive (SMOS, AMSR2) microwave time series data. Signals of these different sensors have individual system-related characteristics (e.g. due to frequency, polarization, or acquisition geometry), but they observe the same media on ground. When considering a time-series of quasi-coincident active and passive microwave signals, they co-vary depending on their sensitivity to the physical properties of the observed media. Such %Based on such characteristic 
covariation patterns allow us consider them jointly to analyze media properties like soil moisture \cite{Jagdhuber2019}, and also, as we show here, to gain understanding of the physical processes governing its variability.   
In the second case study, we consider times series of LAI and fAPAR from MODIS observations, two key essential variables for monitoring terrestrial ecosystems. They are intrinsically related and jointly covary together. Preliminary results of the use of LFMs in remote sensing were given in \cite{Luengo16mlsp} for crop monitoring from optical remote sensing data. Being an optical sensor the presence of clouds can severely hamper parameter estimation. We show how the model fills in the missing data by importing information across different sites, provides credible confidence intervals for the estimates, and helps in recovering the LAI-fAPAR relations from data. This not only allows us to assess the plausibility of the predictions but also to interpret what the GP model has learned in a straightforward way. 

The process convolution model applied in this work is a multioutput GP that shows important advantages over other multioutput GP models, see next section \ref{sec:methods} and Table~\ref{tab:framework}. Remarkably, the model uses a kernel function that emerges from the ODE assumed to govern the system. This endows the model with a high degree of interpretability because, after optimization, one not only learns a predictive model but also the latent forces driving the observations and the associated differential equation parameters which have physical meaning. For instance, the LFM allows us to automatically estimate reasonable parameters of the ODE governing soil moisture dynamics as well as to discover latent forces related to precipitation, see section \ref{sec:experiments}. 
Also, the LFM learns the relationships among the different time series involved in the problem and can `import' information from one to another. In the first case study, for instance, the LFM uses time series of different microwave sensors and allows us to fill in the existing data gaps in an optimal way by transferring information across time series, cf. section \ref{sec:exp2}. In the second case study, this transferability effect is used across space and time as observational time series for LAI and fAPAR are acquired at different sites, cf. section \ref{sec:exp1}. 
In conclusion, the proposed hybrid methodology reconciles mechanistic and data-driven approaches for remote sensing parameter estimation, cf. section \ref{sec:conclusions}, leading not only to accurate but also informative, transparent and interpretable machine learning models for Earth observation.

\section{Methods} \label{sec:methods}

In this section we review the family of process convolution models, widely used in geostatistics and machine learning. 
We explain the most elementary model of the Gaussian process and how various GPs can be mixed, either instantaneously or convolutionally, in order to jointly model multiple signals. The LFM is a convolution process with appealing properties compared to other approaches in terms of interpretability as we will see.

\subsection{Gaussian process regression}

GPs are state-of-the-art tools for regression and function approximation, and have been recently shown to excel in biophysical variable retrieval~\cite{CampsValls16grsm} and model emulation~\cite{Verrelst12,svendsen2020active}. Many applications of plain GP models can be found in the remote sensing literature following either a direct statistical approach in which an inversion GP model is learned from in situ data~\cite{Verrelst12,CampsValls16grsm}, by inverting simulations from a radiative transfer model (RTM)~\cite{CamposTaberner2015,CamposTaberner2016b}, or by fusing both simulations and observational data in the the prior description~\cite{svendsen2017joint}. 
We first fix the notation and briefly review the standard formulation of GPs, and then present the LFM that synergistically combines a mechanistic model, in the form of an ODE, with a data-driven GP approach.

\subsubsection{Model}
Let us consider a set of $N$ pairs of observations or measurements, ${\mathcal D}:=\{t_i,y_i\}_{i=1}^N$, where the inputs $t_i\in\Real$ are the time at which a measurement is acquired\footnote{The input could in general be of higher dimension than one, e.g. it could correspond to the spectral signature recorded by a satellite, but for the purposes of this paper only time is considered as an input.}, and the corresponding outputs $y_i$ represent the parameter of interest (e.g. LAI). In this work we consider time series produced using %several different
a variety of remote sensing sensors.

We assume the following additive noise model,
\begin{equation}\label{GLR}
y_i = f(t_i) + e_i,~~~~~~e_i \sim\Normal(0,\sigma^2),
\end{equation}
where $f(t)$ is an unknown latent function that we aim to learn (which is typically non-linear and non-parametric) that takes time $t\in\Real$ and returns parameter estimates $\hat y$, and $\sigma^2$ stands for the noise variance assumed to be additive and Gaussian.  
Defining $\y=[y_1, \ldots ,y_n]^\intercal$ and ${\bf f}=[f(t_1),\ldots , f(t_n)]^\intercal$, the conditional distribution of $\y$ given ${\bf f}$ becomes $p(\y | {\bf f}) = \mathcal{N}({\bf f}, \sigma^2\I)$, 
where $\I$ is the $n\times n$ identity matrix. Now, in the GP approach, we assume that ${\bf f}$ follows an $n$-dimensional Gaussian distribution ${\bf f} \sim \mathcal{N}({\bf 0},{ {\bf K}})$ \cite{bishop2006pattern}.

The covariance matrix ${ {\bf K}}$ of this distribution is determined by a kernel function encoding the similarity between the input points \cite{Rasmussen06}. Very often the Radial Basis Function (RBF) kernel is adopted, $[\bK]_{ij}:=k(t_i,t_j)=\exp(-\|t_i-t_j\|^2/(2\lambda^2))$, where $\lambda$ is the kernel lengthscale. The intuition here is the following: the more similar input $t_i$ and $t_j$ are the more correlated output $y_i$ and $y_j$ should be. The RBF kernel generally performs well in practice, has good theoretical properties and only involves fitting one hyperparameter $\lambda$. 

\subsubsection{Predictive distribution}
Since the prior and the likelihood are both Gaussian, the marginal distribution of $\y$ can be derived easily
\begin{equation}\label{eq:evidence}
p(\y) = \int p(\y | {\bf f}) p({\bf f}) d {\bf f} = \mathcal{N}({\bf0},{ {\bf K}} + \sigma^2\I).
\end{equation}
Now, we are interested in predicting a new output $y_\ast$ given an input $t_\ast$. The GP framework handles this by constructing a joint distribution over the training and test points,
\begin{equation}
\begin{bmatrix}
\y \\
y_\ast
\end{bmatrix}
\sim
\mathcal{N} \left( {\bf0},
\begin{bmatrix}
{ {\bf K}} + \sigma^2\I & \bk_\ast^\intercal \\
\bk_\ast & k(t_*,t_\ast) + \sigma^2
\end{bmatrix} \right),  \nonumber 
\end{equation}
where $\bk_{*} = [k(t_*,t_1), \ldots, k(t_*,t_n)]^\intercal$ is an $n\times 1$ vector. Then, using the well known properties of Gaussians, we can find the distribution over $y_\ast$ conditioned on the training data, i.e.,
$$
p(y^*|\y)=  \mathcal{N}(\mu_{\text{GP}} (t_\ast),\sigma^2_{\text{GP}} (t_\ast)),
$$
which is a Gaussian distribution with predictive mean and variance given by
 \begin{align} \label{eq:preddist}
  \begin{aligned}
   \mu_{\text{GP}} (t_\ast) &= \bk_{*}^\intercal {({\bf K} + \sigma^2\I)^{-1}}\y , \\ %=  \bk_{*}^\intercal (\bK + \sigma^2\I_n)^{-1}\y, \\
   \sigma^2_{\text{GP}} (t_\ast) &= \sigma^2+ k(t_*,t_\ast)  - \bk_{*}^\intercal ({\bf K} + \sigma^2\I)^{-1} \bk_{*}. % = c_\ast - \bk_{*}^\intercal (\bK + \sigma^2\I_n)^{-1} \bk_{*}.
  \end{aligned}
 \end{align}
% where {\color{red}${\bf K}_{yy}={\bf K} + \sigma^2\I$.}
%Namely, $p(y^*|\y)=\mathcal{N}(y^*|\mu_{\text{GP}}(t_\ast),\sigma^2_{\text{GP}}(t_\ast))$. 
Moreover, denoting $f^*=f(t_\ast)$ we have that $p(f^*|\y)=\mathcal{N}(y^*|\mu_{\text{GP}}(t_\ast),\sigma^2_{\text{GP}}(t_\ast)-\sigma^2)$. Note that the two previous densities only differ in the expression of the variance.

\subsubsection{Learning}

Learning the model hyperparameters  $\thetavec$ can be done by maximizing the marginal likelihood given by Eq.\,\eqref{eq:evidence} with respect to $\thetavec$.
The maximum likelihood (ML) estimator of the hyperparameters is thus obtained by minimizing the negative log-likelihood function
\begin{equation}
	J(\thetavec) =  \vec{y}^{\top} ({\bf K} + \sigma^2\I)^{-1} \vec{y} + \log \vert {\bf K} + \sigma^2\I \vert,
\label{eq:logLikelihood}
\end{equation}
where the dependence on the hyperparameters is contained within $\vec{K}$, $\log$ denotes the natural logarithm, and we have dropped all the elements that do not depend on the hyperparameters.

\subsection{Multi-output Gaussian Processes}
Gaussian process models can be extended to model data from various \textit{sources} or \textit{signals}\footnote{We use the \textit{sources} and \textit{signals} interchangeably in this work.}, exploiting dependencies between them. Such methods are referred to as multi-output Gaussian processes \cite{alvarez2011kernels}, and work by defining a kernel function that distinguishes between data sources.

\subsubsection{Model}
Let us consider a set of $Q$ signals $y_q(t)$, for which we have $N_q$ samples available, taken at sampling points  $t_{q,i}$, i.e. each signal is a set of pairs $\mathcal{D}_q = \{ t_{q,i},\, y_{q,i} \}_{i=1}^{N_q}$ .

In order to jointly model the $Q$ signals we need a \emph{multi-output kernel function} $k_{y_py_q}(t,t')$ 
which computes the covariance between two output-data based 1) on their similarity in input-space as in the case of single-output GPs and 2) on which sources they belong to.
This allows for the added flexibility of assigning e.g. higher covariance between points from the same data source and to determine how much the different signals covary. Thus, when predicting the output of signal $q$ in a new input, the previously observed data involved in the prediction will be weighed according to how similar it is in input-space, but also how similar the signals are. The similarity between signals can be modelled as constant across input-space, as is the case for the well-known {\it linear model of coregionalization} (LMC) \cite{journel1978mining}, or as a function of input-location  \cite{alvarez2011kernels}.

\begin{figure}[!t]
\begin{center}

\vspace{1cm}
\begin{tabular}{c|c}
\hspace{0.5cm}{\bf Independent GPs} \hspace{0.2cm} & \hspace{1.2cm}{\bf Multioutput GP}  \\[3mm]
${\bf K}=\begin{bmatrix}
    {\bf K}_{{\bf f}_1{\bf f}_1} & {\bf 0} \\
    {\bf 0}   & {\bf K}_{{\bf f}_2{\bf f}_2}\\
\end{bmatrix}$ \hspace{0.5cm} 
&
\hspace{0.3cm}  ${\bf K}=\begin{bmatrix}
    {\bf K}_{{\bf f}_1{\bf f}_1} & {\bf K}_{{\bf f}_1{\bf f}_2} \\
    {\bf K}_{{\bf f}_2{\bf f}_1} & {\bf K}_{{\bf f}_2{\bf f}_2} \\
\end{bmatrix}$
\end{tabular}
\caption{
Kernel matrix structure in a multioutput scenario. For simplicity, only two outputs are considered. Developing a different model per output leads to a block diagonal matrix without crossrelations. In a generic multioutput scheme also the cross-covariance matrices are considered, i.e.,  ${\bf K}_{{\bf f}_1{\bf f}_2}$ and ${\bf K}_{{\bf f}_2{\bf f}_1}$. The different multioutput schemes differ by the choice of these cross-terms. In the case of process convolution models, like in LFMs, the kernels are covariance functions derived from the convolution of latent forces and smoothing kernels. Interestingly,  smoothing kernels can be derived from assumed ODEs driving the system, and both smoothing kernels and latent forces can be learned from data.
\label{fig:kernel}}
\end{center}
\end{figure}

\subsubsection{Predictive distribution and hyperparameter learning}\label{sec:multioutputlearning}
The predictive distribution has the same form as for single-output GPs (see Eq.~\eqref{eq:preddist}), where ${\bf K}$ is now a $\sum_{q}^{Q}N_q \times \sum_{q}^{Q}N_q$ matrix. The blocks in the diagonal encode the covariances of points within the same signal, and the off-diagonal blocks hold the covariances between points belonging to different signals. Furthermore, the vector $\vec{y}$ holds the concatenations of observations for each signal $\vec{y}=[\vec{y}_1,\ \ldots,\ \vec{y}_Q]^{\top}$ where $\vec{y}_q=[y_{q,1},\ \ldots,\ y_{q, N_q}]$, and $\bk_{*}$ holds the kernel similarity of the test-point with all the training-points according to the multi-output kernel. With these definitions one can use Eq.~\eqref{eq:logLikelihood} to fit the hyperparameters.\\
Note that if one were to model each signal as an independent GP, it would correspond to have 0's everywhere in a multioutput kernel matrix except the block-diagonal which describes the covariance of points within each individual signal - see Figure \ref{fig:kernel} for illustration.

\subsubsection{Instantaneous and convolutional mixing}
A simple way to construct a multi-output kernel function is the linear model of coregionalization (LMC), also known as \emph{co-kriging} in the field of geostatistics~\cite{Journel78}. It models the dependence between $Q$ outputs by assuming that each output is a linear combination of $R$ independent Gaussian process functions $y_q(t) = \sum^R_r a_{qr} f_r(t)$. This results in a \textit{separable} multi-output kernel function $k_{q,p}(t,t') = \sum^R_r a_{q,r}a_{p,r} k_r(t,t')$, in the sense that it is a product of functions that encode the dependence between outputs $k_r(q,p) = a_{q,r}a_{p,r}$ and functions that describe similarity in input space $k_r(t,t')$. The LMC model is an example of \textit{instantaneous mixing} of Gaussian processes, since each $y_q(t)$ depends on the underlying functions $f_r(t)$ exactly at time $t$.

\begin{figure}[!t]
\begin{center}
\includegraphics[width=7cm]{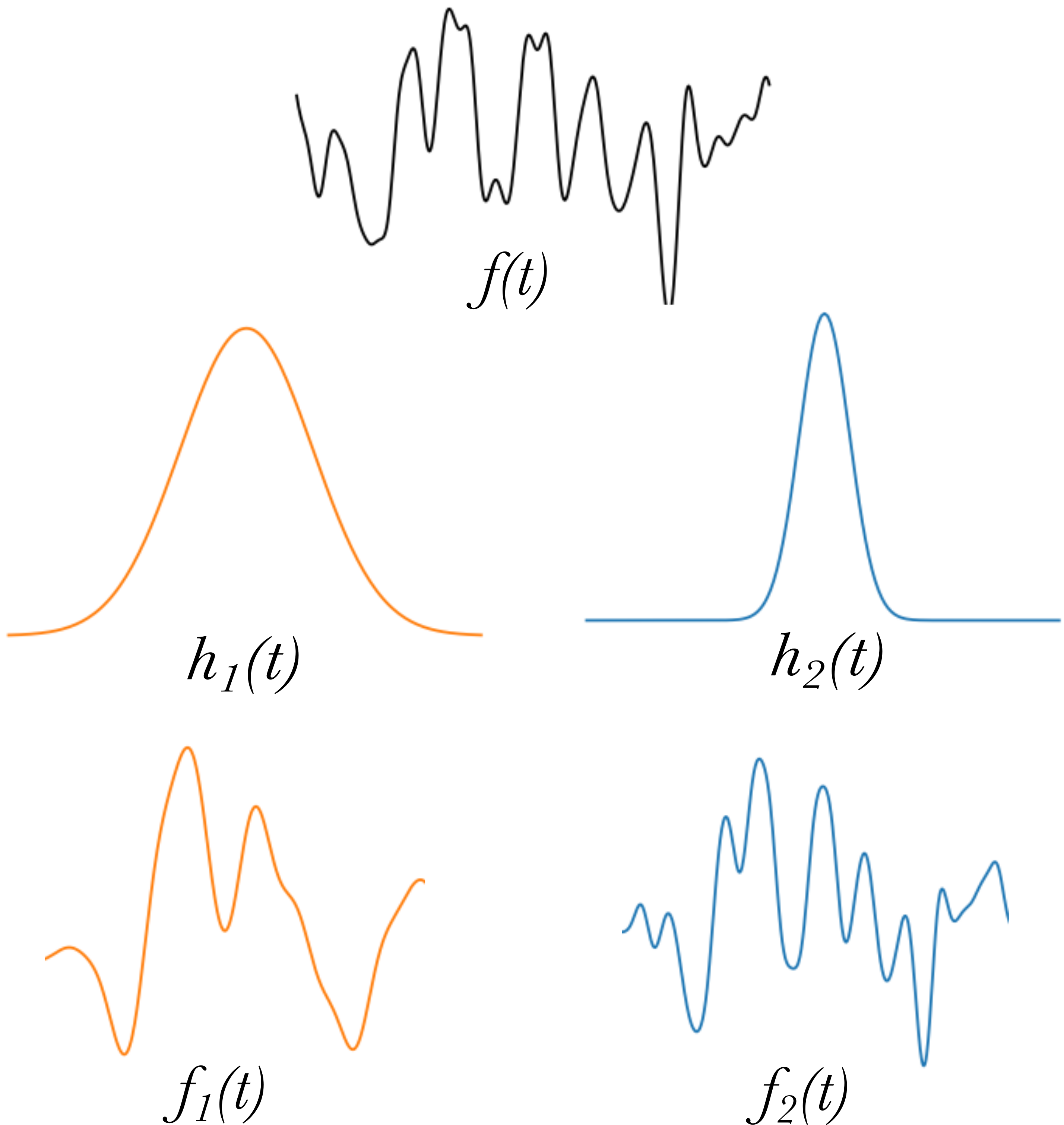}
\vspace{0.2cm}
\caption{
A process convolution illustration with two output functions ${\bf f}_1$ and ${\bf f}_2$. Each function can be expressed through a convolution integral between a (smoothing) kernel $h_1(t)$ and $h_2(t)$ and a latent function (or force) $f(t)$, $f_q=\int_{\mathcal T}h_q(t-\tau)f(\tau)\text{d}\tau$, $q=1,2$.  %tentative figure\label{fig:processconvo}... GCV: the second one is from Mauricio
\label{fig:processconv}}
\end{center}
\end{figure}

A more flexible way of constructing multi-output kernels comes about by assuming that the $Q$ signals we are modelling are generated by a convolution between $Q$ different \textit{smoothing kernels} and a latent Gaussian process $f$:
\begin{equation}\label{eq:processconvolution}
y_q(t)=\int_{\mathcal{T}} h_q(t-\tau) f(\tau) \diff\tau.    
\end{equation}
This generative model explains the correlation between signals through the various convolutions with the same underlying function $f$. This expression is a linear operator working on a Gaussian process which implies that $y_q(t)$ itself is a Gaussian process~\cite{bishop2006pattern}. This model is an example of \textit{convolutional mixing}, as each $y_q(t)$ depends on the underlying GP function at a range of different input values. The model easily extends to include more than one latent function under the assumption that they are independent \cite{alvarez2011computationally}. Denoting the kernel function of $f$ as $k_{f}(t,t') = \mbox{cov}(f(t),f(t'))$, the outputs will be governed by a Gaussian process with a multi-output kernel function: 
\begin{align}
	k_{y_p y_q}&(t,t') = \mbox{cov} ( y_p(t),\, y_q(t') ) \nonumber \\
	&=  \mbox{cov} \left( \int_{\mathcal{T}} h_p(t-\tau) f(\tau) \diff\tau, \int_{\mathcal{T}} h_q(t'-\tau') f(\tau') \diff\tau' \right) \nonumber \\
	&= \int_{\mathcal{T}} h_p(t-\tau) \int_{\mathcal{T}} h_q(t'-\tau') k_{f}(\tau,\tau') \diff\tau' \diff\tau.
\label{eq:processconvolutionkernel}
\end{align}
For this double integral to exist the smoothing kernels need to have compact support or be square-integrable \cite{alvarez2011kernels}. Depending on the form of the smoothing kernels and the assumed number of latent functions, an arbitrarily flexible multi-output kernel can be constructed. See Fig.~\ref{fig:processconv} for a simple illustration of process convolutions with $Q=2$. If domain knowledge of the specific problems exists, however, it is often desirable to build a physics-based interpretable kernel function.
In the following we describe a process-convolution kernel which corresponds to the solution of a linear system of differential equations.

\begin{table*}[t!]
 \caption{Characterization of the discussed GP models in this paper and used in geostatistics and machine learning.% and based on mixing processes.}%, both instantaneous and convolutional.%\red{GCV: 1) maybe later, after presenting proc.conv. as a summary? and also 2) we need to cite it properly and several times?}
 }
 \small{
    \label{tab:framework}
    \centering
    \begin{tabular}{lllll}
    \hline
Model & Form of smoothing kernel & Form of kernel &  Interpretability & Multioutput \\
    \hline
%KRR \cite{ShaweTaylor2004} & $\delta_i$ & {\bf K} & Instantaneous & L & $\times$ & $\times$  & $\times$ \\
Ind. GPs \cite{Rasmussen06} & Delta function & $\sum_{r}^Q \delta_{qr}k_r(t,t')$   &  Kernel dependent & $\times$  \\
LMC \cite{Goovaerts97} & Delta function & $\sum_{r}^R b_{qp}k_r(t,t')$   & Kernel dependent & $\surd$ \\
\hline
PC \cite{ver1998constructing} & Square-integrable  & $\sum_{r}^R \mbox{cov}(\,(h_q \ast f_r)(t), (h_p \ast f_r)(t')\,)$    & Kernel dependent & $\surd$ \\
LFM \cite{alvarez2013linear} & Green's function & $\sum_{r}^R \mbox{cov}(\,(G_q \ast f_r)(t), (G_p \ast f_r)(t')\,)$  & High: ODE params, LFs & $\surd$ \\
    \hline
    \end{tabular}
    }
\end{table*}

\subsection{Gaussian Process Latent Force Models} % \addan{for soil moisture?}} 

We turn now to the interesting scenario where we assume that the signals $y_q(t)$ are governed by an underlying ordinary differential equation (ODE) with a Gaussian process latent force. We follow the procedure described in \cite{alvarez2009latent} which shows that the solution to the ODE is itself a Gaussian process over the outputs with a multi-output kernel which contains parameters of the underlying ODE.

\subsubsection{Model}
%\yy{Plan: Start from n'th order ODE and state that it can be solved with greens functions form }

Once again we consider a set of $Q$ signals $\mathcal{D}_q = \{  t_{q,i},\, y_{q,i} \}_{i=1}^{N_q}$.
We assume that the signals are governed by a system of ordinary differential equations (ODEs) defined by the linear differential operators $L_q$, and acted on by $R$ latent force functions:
\begin{equation}\label{eq:ode}
    L_q\{y_q(t)\} = \sum_{r=1}^R S_{r,q} f_r(t).
\end{equation}
The correlation between signals thus arises as a result of their coupling with the set of $R$ latent forces (LFs). Under null initial conditions, the solution $y_q(t)$ to a linear differential equation with constant coefficients can be expressed as the convolution of the latent force functions (right hand side of Eq. \eqref{eq:ode}), with the impulse response function $G_q(t)$ of the differential operator which encodes all the information of the ODE:
\begin{equation}
	y_q(t) = \sum_{r=1}^{R}{S_{r,q} \int_{0}^{t}{f_r(\tau) G_q(t-\tau) \textrm{d}\tau}}.
%\label{eq:outputsSpecific}
\end{equation}
This, we can immediately recognize as a process convolution, where the smoothing kernel takes the form of the impulse response function $G_q(t)$. The outputs $y_q(t)$ are therefore governed by a Gaussian process with a multi-output kernel given by Eq. \eqref{eq:processconvolutionkernel}. Assuming that the latent forces  are independent we get 
\begin{align} \label{eq:lfmkernel}
	k_{y_p y_q}(t,t') =& \sum_{r=1}^{R} S_{r,p} S_{r,q} \int_0^t G_p(t-\tau) \int_0^t G_q(t'-\tau') \times \nonumber \\ &k_{f_r}(\tau,\tau') \diff\tau' \diff\tau.
\end{align}
where $k_{f_r}$ is the kernel function of the Gaussian process prior over latent force $f_r(t)$. The form of the multi-output kernel depends on the impulse response of the differential operator and will therefore contain physically meaningful parameters of the underlying ODE. For some differential operators and choices of priors over latent forces we obtain a closed-form solution, as will be shown below.

This framework also allows us to infer the latent forces which act upon the physical system. In order to derive their predictive distribution we need to compute the covariance between observations and latent forces $\mbox{cov}(f_r(t),\, y_q(t'))$. The cross covariance kernel is given by
\begin{equation}
k_{f_r y_q}(t,t') = S_{r,q} \int_0^t k_{f_r}(t,\tau) G_q(t'-\tau)  \diff\tau
\end{equation}
which is a simpler expression to compute than the double integral of the multi-output kernel above.
Table \ref{tab:framework} provides a brief summary of the different discussed approaches.

\subsubsection{First-order differential equation}

A useful model that recurrently appears in many fields of science and in remote sensing problems in particular is a first order ODE. As we will see in the experimental section, this simple first-order ODE is quite efficient in capturing the exponential decay behaviour exhibited in soil moisture data. Let us consider a {non-homogenous} ODE of the form
\begin{equation}
    \frac{\mbox{d}y_q(t)}{\mbox{d}t} + D_q y_q(t) = B_q + \sum_{r=1}^R S_{rq} f_r (t),\label{eq:ODE}
\end{equation}
where $f_r$ is a latent force on the system. In the example of modelling soil moisture (SM) dynamics with Eq. \eqref{eq:ODE}, the latent force could be related to rainfall adding moisture or radiation evaporating it, parameter $B_q$ will be mainly related to soil hydraulic properties, and parameter $D_q$ will represent the decay rate \cite{McColl2017}. The inverse of the decay rate is the $\tau$ or \textit{e-folding time} parameter, which is typically used as a measure of soil moisture persistence \cite{Delworth1988}.

As described before, Eq. \eqref{eq:ODE} can be solved as the convolution of the latent force with the impulse response of the ODE \cite{arfken1999mathematical}, to obtain
\begin{equation} \label{eq:sol}
    y_q(t) = \frac{B_q}{D_q} + \sum_{r=1}^R S_{rq} \exp(-D_q t) \int_0^t f_r(\tau ) \exp (D_q \tau ) \mbox{d} \tau .
\end{equation}
As previously described, our assumptions imply that $y_q(t)$ is governed by a Gaussian process with a multi-output kernel $k_{y_p y_q}(t,t') = \mbox{cov}[y_p(t),y_q(t')]$.
Inserting \eqref{eq:sol} into the covariance expression, one can arrive at a closed-form solution under the assumption that each latent force GP is independent and has an RBF covariance function \cite{lawrence2007modelling}. The resulting multi-output kernel is given by
\begin{equation}
    k_{y_p y_q}(t,t') = \sum^R_{r=1} \frac{S_{rp}S_{rq} \sqrt{\pi}l_r}{2} [ h_{qp}(t',t) + h_{pq}(t,t') ]
\end{equation}
where
\begin{alignat}{2}
    h_{pq}(t',t) &= \frac{\exp (\nu_{rq})}{D_p + D_q} \exp (-D_qt') \biggl \{\exp (D_qt) \nonumber \\
    &\times  \left[  \mbox{erf} \left( \frac{t'-t}{l_r} - \nu_{rq} \right) +  \mbox{erf} \left( \frac{t}{l_r} + \nu_{rq} \right) \right] \nonumber \\
&- \exp(-D_pt) \left[  \mbox{erf} \left( \frac{t'}{l_r} - \nu_{rq} \right) +  \mbox{erf} \left(  \nu_{rq} \right) \right] \biggr \}
\end{alignat}
in which erf($x$)$=\frac{2}{\pi}\int_0^{x}{\exp{-y^2}dy}$ is the real valued error function and $\nu_{rq} = l_rD_q/2$.

With a multi-ouput kernel encoding the covariance between all the observations from different sources we can write up the log marginal likelihood of all the data as described in \ref{sec:multioutputlearning} and optimize it with respect to kernel hyperparameters. Automatic differentiation is employed using the python package \href{https://github.com/HIPS/autograd}{autograd} and optimization is performed with the ADAM algorithm \cite{kingma2014adam}, which have good and stable default settings and the learning rate is automatically set. This is very interesting because it means we can perform inference about the parameters of the underlying system of differential equations and thus learn from data meaningful physical parameters describing the system, such as the scalings and decay rates of the different signals\footnote{For implementation see \href{https://github.com/dhsvendsen/LFM4RS/}{github.com/dhsvendsen/LFM4RS}.}. Furthermore, we can also derive an expression for the cross-covariance between latent forces and the outputs:
\begin{alignat}{2}
    k_{y_q f_r}(t,t') &= \frac{S_{rq}\sqrt{\pi} l_r}{2} \exp(\nu_{rq}^2)\exp(-D_q(t-t')) \nonumber \\ 
&\times \left[  \mbox{erf} \left( \frac{t'-t}{l_r} - \nu_{rq} \right) +  \mbox{erf} \left( \frac{t'}{l_r} + \nu_{rq} \right) \right]. \label{eq:crosscov}
\end{alignat}
This means that we can perform prediction on and visualize the latent forces which, as we will see later, have a clear physical interpretation. Supplementary material comparing the hybrid approach to purely mechanistic and purely data-driven ones can be found at \href{https://isp.uv.es/lfm/}{isp.uv.es/lfm}.

\subsubsection{On the impulse response}

The impulse response $G_q(t)$ plays the role of the smoothing kernel of process convolutions, and encodes our domain knowledge about the ODE system. Several orders of the ODEs are possible here. In the experimental section, we show how the LFM framework successfully models the soil moisture dynamics by using an impulse response corresponding to a first-order ODE. This is a reasonable choice as it models the exponential decay in time of SM, as argued before. 
A second order ODE could imply assuming a damped sinusoidal kernel, which has been previously used in motion capture data \cite{alvarez2009latent}, and that exhibits more complex nonlinear dynamics related to inertia and resonance. 
Our second experiment deals with the estimation of LAI and FAPAR in time, which exhibit a strong seasonal cycle. In this case, there is no clear governing equation available in the remote sensing literature that one could include in the LFM. In this case, we suggest to use a smoothing kernel that in the process convolution literature has shown good results for similar signals $G_q(t) \propto \exp(-\frac{t^2}{2 \nu_q^2})$. This kernel actually corresponds to assuming an impulse response (Green's function) of the higher order (higher than 1)  differential  equations for modeling (see, for instance, \cite{Luengo16mlsp}). 
This is consistent with the intrinsic relation of the different physiological processes involved in plant growth and development, which are captured by LAI and FAPAR time series.

\section{Experimental results} \label{sec:experiments}

This section provides empirical evidence of the performance of the proposed GP latent force model in two different settings. First we deal with soil moisture time series modelling and gap filling using multiple microwave satellite products. We focus on the acquisitions over an in-situ network equipped with ground measurements of soil moisture and precipitation. Here we assume a first order ODE -the surface water balance equation- that %captures the exponential decay behaviour exhibited in SM data
describes SM changes in time and learn the parameters and unobserved latent processes encoded in the observations. The second problem deals with a multiple output gap filling problem involving LAI and fAPAR times in two distinct sites. Here we have limited knowledge about the dynamic model governing the system, and thus rely on a flexible latent force coupling mechanism (or smoothing kernel). % based on common sense and understanding of plant physiological growth processes. 
In both problems we explore the LFM capabilities to perform gap filling in multisensor and multisite settings respectively, %assessing the prediction accuracy and robustness to high missing data rate regimes, 
and study the learned latent forces and parameters inferred to gain insight in the problem.

\begin{figure*}[t!]
    \centering
    \includegraphics[width=0.75\textwidth]{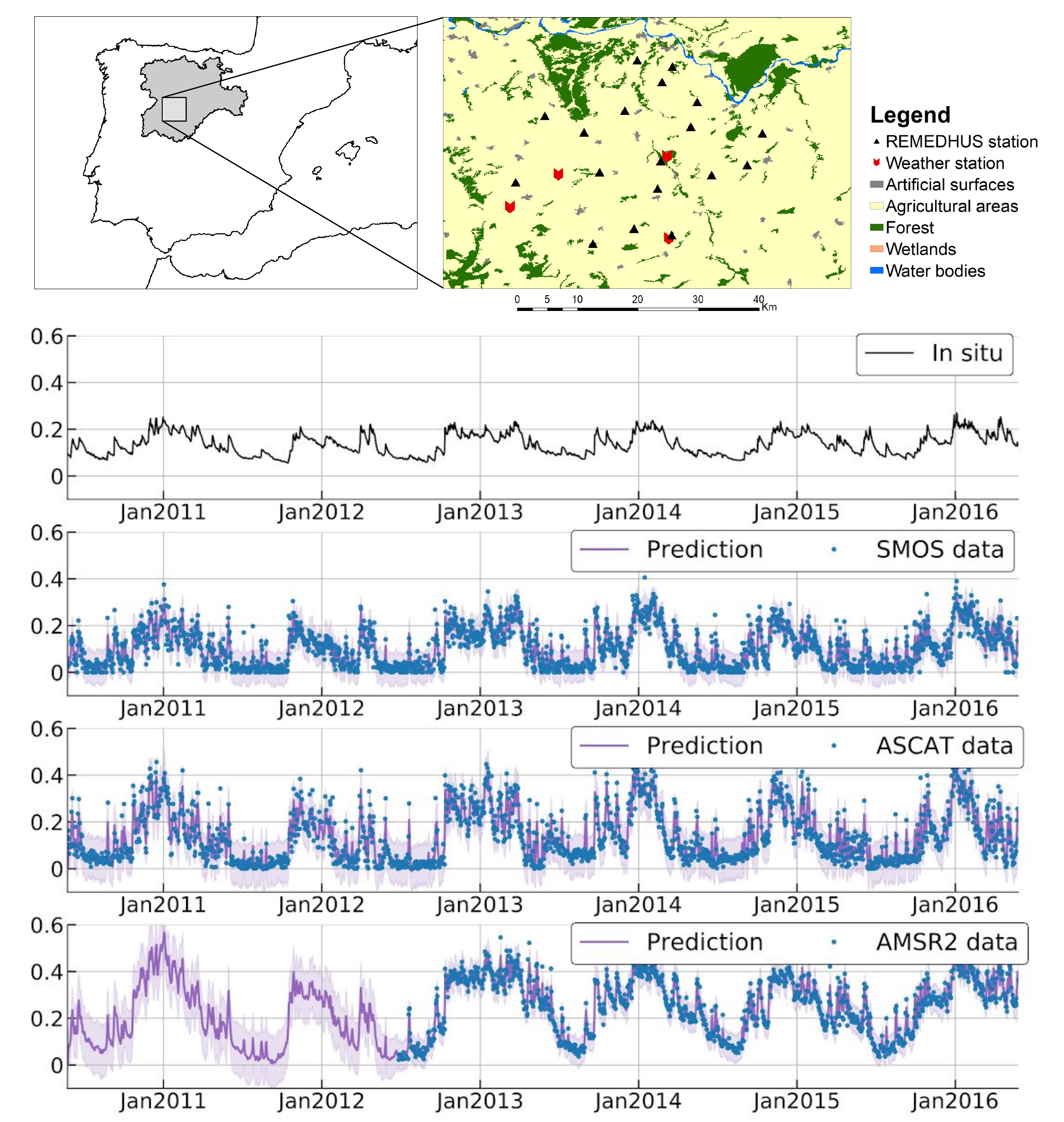}
    \caption{Results of application of multi-output process convolution to soil moisture time series at the REMEDHUS network using three latent forces. Top: layout of the 18 selected soil moisture stations and the 4 weather stations within the REMEDHUS validation site, located at the central part of the Duero river basin, Spain. Bottom: time series of \textit{in-situ} (average of 18 stations), and satellite-based soil moisture estimates (m$^3\cdot$m$^{-3}$) from SMOS, ASCAT and AMSR2 (blue dots denote the training data and purple lines and shaded regions represent the LFM predictions and confidence intervals).} 
    \label{fig:lfmpred}
\end{figure*}

\begin{table*}[t!]
    \caption{Results of application of multi-output process convolution to satellite-based soil moisture time series over the REMEDHUS network using 1, 2, and 3 latent forces. The estimated input noise $\sigma$ and e-folding time $\tau$ (days) obtained per each satellite are reported. The latent force which is more predictive of precipitation in each case is used to calculate: i) the Pearson correlation $R$ of the obtained LF with in-situ precipitation measurements, and ii) the area under the curve ($AUC$) performance metric for classification of rain-events, in which a measured in-situ precipitation higher than 1~mm is considered a rain event (see Fig.~\ref{fig:auc} for more results).}
    \label{tab:sm_results}
\begin{center}
    \begin{tabular}{|l|l|l|l|l|l|l|l|l|l|l|l|l|}
\hline
\multirow{2}{*}{} & \multicolumn{4}{l|}{1 latent function} & \multicolumn{4}{l|}{2 latent functions} & \multicolumn{4}{l|}{3 latent functions} \\ \cline{2-13} 
                     & $\sigma$ & $\tau$ & $R$ & $AUC$ & $\sigma$ & $\tau$ & $R$ & $AUC$ & $\sigma$ & $\tau$ & $R$ & $AUC$ \\ \hline
                 SMOS &   $1.14\times 10^{-3}$    &   67.18  &  \multirow{3}{*}{}  & \multirow{3}{*}{} &  $1.15\times 10^{-3}$   &   23.27 &  \multirow{3}{*}{} & \multirow{3}{*}{}  &   $0.89\times 10^{-3}$   &    16.21 &  \multirow{3}{*}{} & \multirow{3}{*}{}  \\ \cline{1-3} \cline{6-7} \cline{10-11}
                 ASCAT & $3.61\times 10^{-3}$  & 69.54  &   0.517 & 0.811  &  $1.51\times 10^{-3}$ &  19.79   &   0.521  &  0.829  &  $1.26\times 10^{-3}$  &  11.04  &  0.549 & 0.851\\ \cline{1-3} \cline{6-7} \cline{10-11}
                 AMSR2 & $1.40\times 10^{-3}$  & 125.71   &    &  &   $0.44\times 10^{-3}$  & 47.79  &  &    & $0.51\times 10^{-3}$   &  16.27   &   &   \\ \hline
\end{tabular}
\end{center}
\end{table*}

\subsection{Experiment 1: Learning soil moisture dynamics from multisensor data}
\label{sec:exp2}

In the following, we present a soil moisture (SM) prediction application of a GP model which is \textit{derived directly from the governing equations} of the problem at hand. This implies strong model assumptions, but also results in model hyperparameters that have a clear physical interpretation, which is a strong advantage of this type of model.

\subsubsection{Data collection}

Soil moisture is a key hydrologic state variable, important to the understanding of various climatologic and meteorological processes \cite{babaeian2019ground}. The monitoring of SM has many applications such as drought prediction, agricultural yield prediction, forest-fire and flood prevention to name a few. We considered soil moisture time series from three spaceborne microwave sensors, which are integrated in the Copernicus Climate  Change Service (C3S) long-term SM product \cite{Dorigo2017}: the L-band  radiometer on ESA's Soil Moisture and Ocean Salinity (SMOS),  JAXA’s  C-band  Advanced Microwave  Scanning  Radiometer-2  (AMSR2),  and  Eumetsat's  C-band  Advanced  Scatterometer  (ASCAT). The experiment was conducted at the REMEDHUS soil moisture measurement network\cite{Sanchez2012}, located in the central part of the Duero basin (41.1-41.5$^\circ$ N, 5.1-5.7$^\circ$ W), Spain (see Fig.\ref{fig:lfmpred}). It covers  a  semi-arid continental-Mediterranean agricultural region of 35 x 35 km, and has extensively been used for calibration and validation of soil moisture products (e.g.\cite{Sanchez2012,Polcher2016,GonzalezZamora2019}). %The period of study was six years, starting in June 2010. 
Satellite daily soil moisture products (SMOS BEC L3 v.2 \cite{BEC2019}, AMSR2 ESA CCI Passive v3.3.\cite{Dorigo2017}, and ASCAT ESA CCI Active v.3.2 \cite{Dorigo2017}) were projected to a common grid (EASE2 25 km) and used in this study. Ground-based measurements of top 5 cm soil moisture (average of 18 stations) and precipitation (average of 4 automatic weather stations) from REMEDHUS were used as supporting information to interpret and validate the learned latent forces and parameters of the surface water balance equation.

\subsubsection{LFM for multisensor fusion and gap filling}

We considered six years of SM estimates from the three satellites, starting in June 2010. The AMSR2 satellite was launched on May 18, 2012 and hence has no coverage before this date, as shown in Fig.\ref{fig:lfmpred}. Furthermore, the SMOS, ASCAT and AMSR2 time series have an average missing data rate of 8.6$\%$, 23.7$\%$ and 26.8$\%$ (and a maximum gap lenth of 6, 5 and 5 days) respectively. We model the signals of the three satellites simultaneously as a multioutput LFM GP. We thus learn the covariances not only between function values of the same signal, but also between signals. This makes them extremely useful for gap-filling (e.g. \cite{alvarez2011kernels,Pipia2019,Salcedo2020}). Fig. \ref{fig:lfmpred} shows the results of fitting an LFM with 3 latent functions to the SMOS, ASCAT and AMSR2 time series over REMEDHUS. We see that the reconstructed soil moisture time series closely follow the original ones, capturing the wetting-up and drying-down events and filling in the missing information. Predictions have associated confidence intervals related to data uncertainty as well as to model uncertainty. Notably, the method allows for consistent reconstruction of the first two years (pre-launch) of AMSR2 by seamlessly transferring the information from the two other sensors. Interestingly, the confidence intervals in this period exhibit a slow increase with the gap length (see Fig.\ref{fig:lfmpred}). This is different to previous studies with multioutput GPs not including a LFM, in which the confidence intervals were strongly related to data availability only and the impact of model uncertainty was minimal \cite{Salcedo2020}. Although the back-propagation of AMSR2 estimates is shown here only for illustration purposes, our analyses suggest the LFM is capable of dealing with both short and long data gaps in multi-satellite records, providing predictions at all time stamps where at least there is one satellite measurement.

\begin{figure*}[htb]
    \centering
    \includegraphics[width=0.87\textwidth]{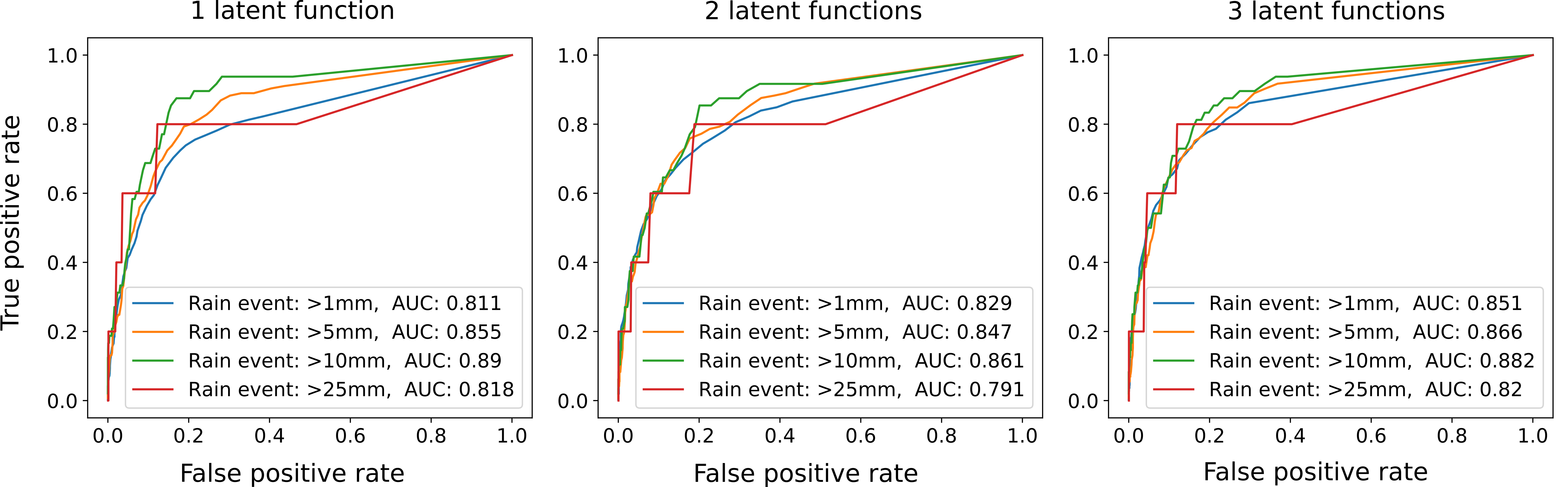}
    \caption{ROC curves for the classification of rain-events using the latent force which is most predictive of precipitation, for each of the three considered LFM models. Four scenarios were considered where a day was labeled as rainy if more than 1, 5, 10 and 25~mm of precipitation was measured. This corresponds to 417, 145, 48 and 5 rainy days respectively in the 6-year study period. We see that the classification performance for rain-events of higher than $1mm{\cdot}day^{-1}$, as measured by area under curve (AUC), increases with model complexity.} 
    \label{fig:auc}
\end{figure*}

\subsubsection{Learning the latent forces of soil moisture}

Using the soil as a rain gauge to estimate precipitation globally via remote sensing measurements of soil moisture from space has shown promise in recent years (e.g. \cite{Brocca2014,Koster2016}). In this experiment, we show how we can naturally obtain precipitation as a latent force of surface soil moisture dynamics using GPs and a simple mechanistic model of the surface water balance (see Eq.~\eqref{eq:ODE}). Physically, the model represents a stochastic (noisy) process possessing some memory and an inherent exponential damping, which is continually forced by a random process. This latent force is naturally that of precipitation, which matches well with latent forces derived from fitting the LFM to the SM time series, as we will show. Note since soil moisture has volumetric units (m$^3\cdot$m$^{-3}$), the depth of the soil volume below the surface or soil sensing depth needs to be added to Eq. \eqref{eq:ODE} as a parameter to fully describe the surface water balance \cite{Delworth1988}. In our experiments, we did not impose any sensing depth and as such the scaling factor of the latent forces are not readily interpretable.

\begin{figure*}[htb]
    \centering
    \includegraphics[width=\textwidth]{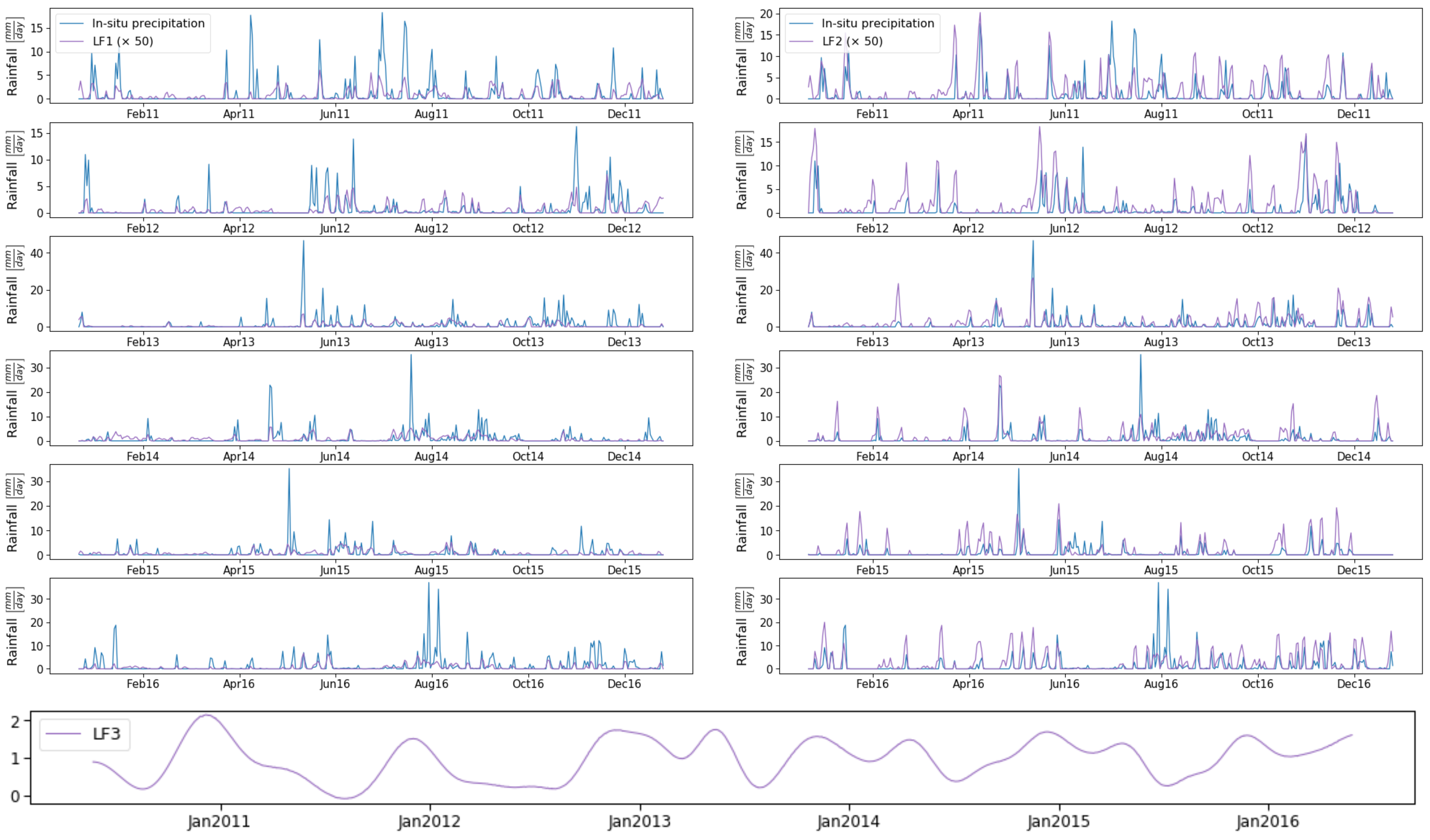} 
    \caption{Inferred latent forces associated to the satellite soil moisture time series over REMEDHUS. Top: first (left) and second (right) estimated latent forces, plotted alongside in-situ precipitation measurements (mm$\cdot$day$^{-1}$). To better illustrate the comparison, the negative values of the LFs are set to zero and a scaling factor is applied. Bottom: third estimated latent force, which may be related to the annual trend or seasonality inherent to Earth system processes.}
    \label{fig:lf23}
\end{figure*}

Using the cross-covariance function between the observations and the (unobserved) latent forces in Eq. \eqref{eq:crosscov}, we can reconstruct the latent forces. %These are plotted in figure \ref{fig:lf23} along with the precipitation measured in-situ at the REMEDHUS site.
These latent forces capture different aspects of the signal, as shown in Fig.~\ref{fig:lf23}. The first latent force exhibits peaks that match well with the in-situ precipitation but has a lengthscale that varies too slowly to capture the spiky nature of the actual rainfall. The second LF essentially captures the narrow peaks of the precipitation but also suffers more false positives, i.e. claims that moisture was added to the ground when no rain was recorded. The third latent force captures the seasonality of the time series. The negative values of the first and second latent forces were set to zero, as done in previous works that attempt to model the precipitation in more direct ways \cite{Brocca2014}. Also, a scaling of 50 was applied so that the LFs lie within the range of in-situ precipitation and we can better compare them. %This scaling, learned indirectly from comparison of LFs with precipitation, can be interpreted as the soil sensing depth of the measurements, which was not implicitly included in the ODE. Remarkably, the obtained scale factor lies within the theoretical sensing depth of the microwave satellites, of about 10 mm at C-band and 50 mm at L-band (units are mm since precipitation is provided in $mm{\cdot}day^{-1}$).
Notably, using three latent forces allowed the model to fit the data better than using one or two both in terms of marginal likelihood and of estimated e-folding time and correlation of inferred LFs with precipitation (See Tab.~\ref{tab:sm_results}). Attempting to fit more than three latent forces did not improve accuracy nor the interpretability of the model, it only added computational complexity and training time.  Unless prior knowledge of the true number of latent forces is available, %Physical interpretability of the latent forces is not ensured and
it is recommended to keep the number of latent forces to the minimum number that still maximizes predictive accuracy.

Precipitation estimation is important to the development of a proper understanding of the hydrologic cycle as water passes through the ocean, land, and atmosphere. It is therefore very interesting that the latent forces obtained from fitting a LFM to the soil moisture data correlate well with the in-situ precipitation. %The LFM has different local minima which makes it difficult to find the exact coupling constant of the latent forces. 
Due to the high presence of zeros (days with no rain) in the precipitation data, the Pearson's correlation coefficient reported in Tab.~\ref{tab:sm_results} is not the best performance metric. We also apply the latent force for the classification problem of predicting rain-events. The area under curve (AUC) score for four different scenarios is shown in Fig.~\ref{fig:auc}. The best prediction capabilities are obtained in terms of correlation coefficient and AUC when the three latent functions are considered in the LFM.

\subsubsection{Inferring physically meaningful hyperparameters}
Alongside precipitation, we infer the e-folding time scale $\tau$, which is typically utilized as a measure of the soil moisture memory or persistence \cite{Koster2001,Seneviratne2006}. This parameter is relevant for climate models, since sufficient soil moisture memory is a necessary condition for the occurrence of soil moisture-precipitation feedbacks \cite{Sellers1997,Koster2003}. 
While here we directly learn the $\tau$ parameter from the decay rate of the ODE, it can also be estimated from the autocorrelation of the soil moisture time series. After removing the seasonal cycle from the time series, $\tau$ corresponds to the time lag at which the autocorrelation decreases to a factor of $e^{-1}\approx0.37$ \cite{Delworth1988,Rebel2012}. Application of this technique to REMEDHUS in-situ time series leads to a $\tau$ of $\sim 16$ days. When applied to SMOS, ASCAT and AMSR2 time series, estimates range from 9 to 11 days \cite{Piles18xcorrsm}, which are close to our estimates with three latent forces (See Table \ref{tab:sm_results}). 

The LFM automatically estimates an input noise associated to each of the time series, which is consistent with the sensor's characteristics (See Table \ref{tab:sm_results}). For the three experiments, the model assigns the highest input noise to the active sensor (ASCAT) and, among the passive sensors, a slightly higher noise is assigned to SMOS than to AMSR2 for the cases of using 2 and 3 LFs. The latter is in agreement with the typically lower signal-to-noise ratio of aperture synthesis radiometers (i.e. SMOS) with respect to real aperture ones (i.e. AMSR2).

To maintain tractability, non-linear relations were not included in the differential equation used in this LFM and perhaps this limited our capability of estimating the parameters. Nonetheless, we showed that by increasing the number of LFs we were able to fit the model better to the data and provide credible estimates of the e-folding time. The use of more complex ODEs could allow the estimation of soil parameters relevant for climate studies and is recommended for further research.

\subsection{Experiment 2: Single sensor, multi-site gap filling}
\label{sec:exp1}

\begin{figure*}[!htb]
	\centering
	\includegraphics[width=0.32\textwidth]{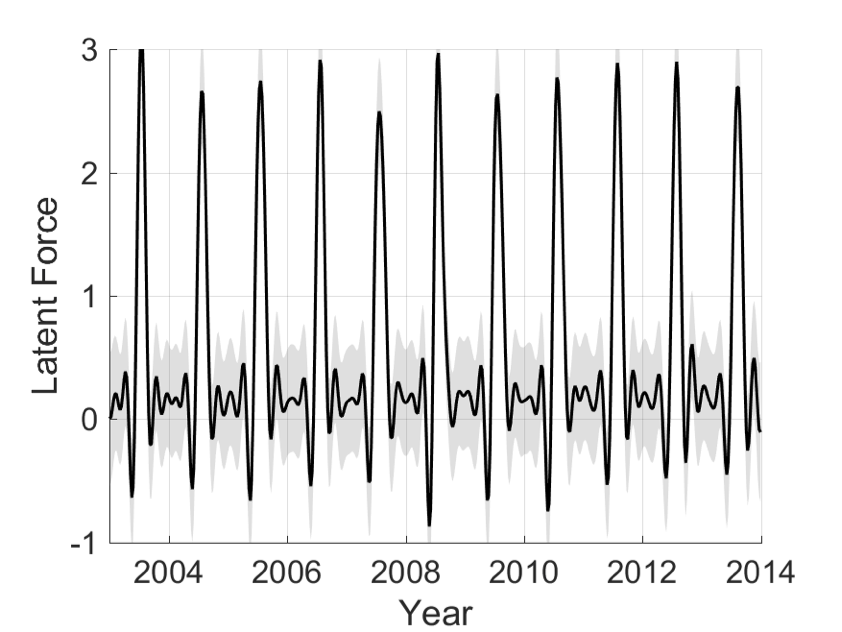}
	\includegraphics[width=0.32\textwidth]{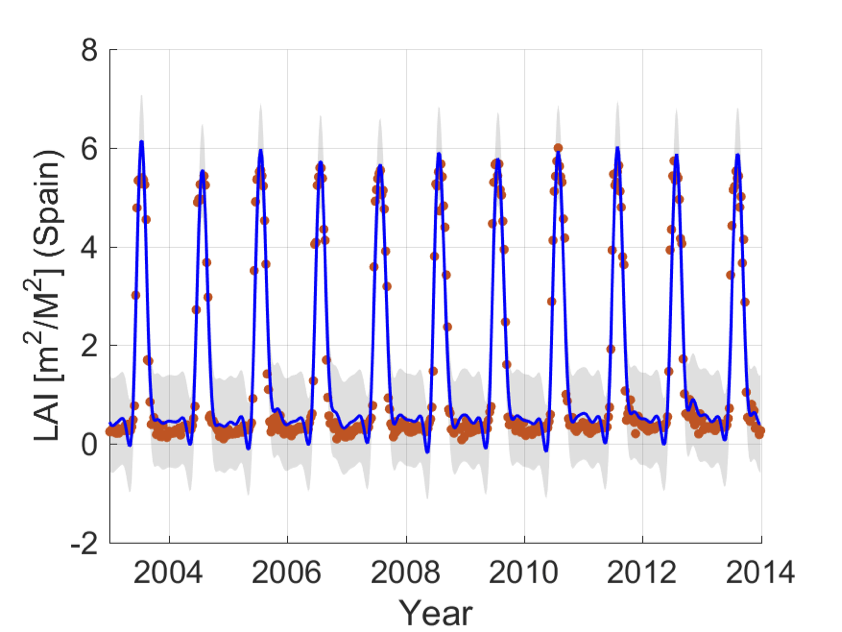}
	\includegraphics[width=0.32\textwidth]{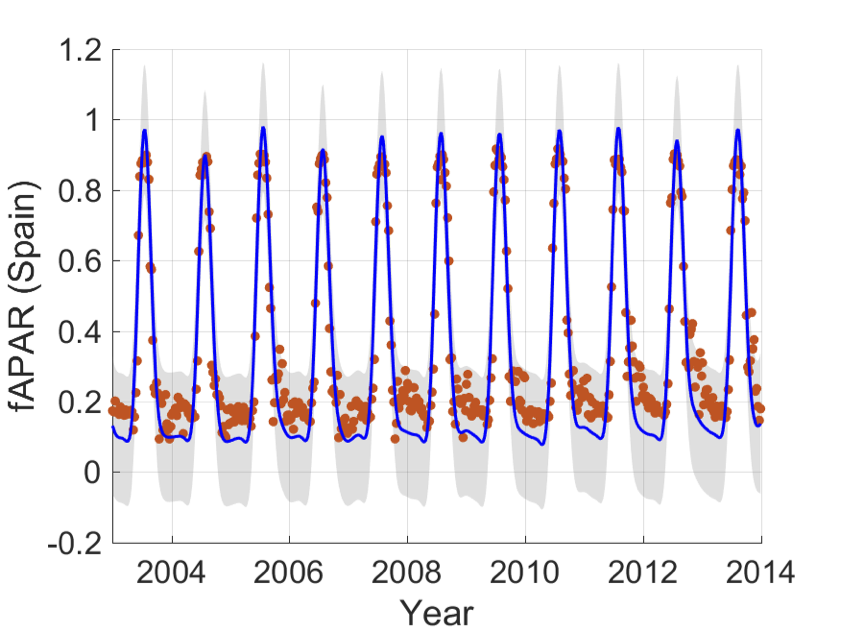}
\caption{Learning a single LF \textbf{(left)} using all the LAI and fAPAR data for Spain and Italy. \textbf{(Middle and right)} LAI and fAPAR for Spain: training data (red dots), predicted time series (blue line) and uncertainty measured by $\pm 2$ standard deviations about the mean predicted value (grey shaded area). }
\label{fig:res_noMissing}
\end{figure*}

\subsubsection{Experimental setup and data collection}

We highlight the usefulness of the proposed method for filling gaps in time series of two of the main remote sensing bio-physical variables such as Leaf Area Index (LAI), which provides valuable information to climate and hydrologic modelling, and the fraction of absorbed photosynthetically active radiation (fAPAR), which is a key variable in the assessment of vegetation productivity and yield estimates. We will use remote sensing time series from different sites and the different variables for multioutput GP modeling and gap filling. 

The applied LFM GP assumes that both the latent force kernels and the smoothing kernels are Gaussian. Since the LFs are zero-mean GPs, the noise is zero-mean and Gaussian, and all the operators involved are linear, the joint LFs-output process is also a GP. 
% The motivation for this assumption is the following.
The choice of a Gaussian smoothing kernel is inspired by the well-known fact that LAI and fAPAR are tightly related. With higher fAPAR, the slope of LAI should be higher (more rapid plant growth rate because of more efficient use of the light incoming radiation) until a plant physiological limit and decline in PAR during the Fall, and thus of fAPAR too. As revised in Sect. 2, assuming a Gaussian smoothing kernel\footnote{When using a Gaussian smoothing  \textit{and} latent force kernel, the resulting convolution process is referred to as a Gaussian-Gaussian.}, $h_q(t)\propto \exp(-\frac{t^2}{2 \nu_q^2})$, implements a Green's function of the heat diffusion equation process. While yielding good results, the hyperparameters fitted in this experiment do not have a clear interpretation as was the case in experiment 1.

In this experiment we use multitemporal LAI and fAPAR products corresponding to the main Mediterranean rice areas. The data were derived in the context of the European \href{http://www.ermes-fp7space.eu/} {ERMES project}, aimed to develop a prototype of COPERNICUS down-stream services based on the assimilation of Earth observation (EO) and \emph{in situ} data within crop modeling solutions dedicated to the rice sector. In this framework, the ERMES study areas have been selected in Spain and Italy, and Greece, which are the three countries responsible of 85\% of total European rice production. The Spanish study area extends from the Ebre river delta area to the \emph{Parc Natural de l'Albufera} in Val\`encia, and is situated on the Spanish Mediterranean coast covering from Tarragona to the region of Val\`encia, and is one the main rice cultivation areas of the country. The Italian study area is located in the Lomellina rice district, which is in the south-western Lombardy region, between the Ticino, Sesia and Po rivers. The Greek study area covers the two main rice cultivation areas of the country, Thessaloniki (180.000 ha) and Serres (4.000 ha), where cultivation represents almost the 75\% of the Greek rice area.
Within each study area, rice is a common crop with a long tradition and cultural and economical value.

For rice, LAI ranges between values close to zero to a maximum of 10 at flowering, although maximum values closer to 6 are the norm, while fAPAR ranges from 0 to 1. In this paper we focus on a set of representative rice pixels of each ERMES area, thus allowing us to observe the inter-annual variability of rice from 2003 to 2014 at coarse spatial resolution (2~Km) which is useful for regional vegetation modelling. Concretely, the ERMES LAI and fAPAR products were derived from two optical sensors onboard satellite platforms (MODIS and SPOT-VGT) thus obtaining similar multitemporal trends computed over different dates from two different sources. The ERMES LAI/fAPAR based on SPOT-VGT data provides estimates every 10 days while the one based on MODIS data provides estimates every 8 days. 

\subsubsection{Multi-output time series: learning the latent forces}

In this section, we focus on learning the latent forces for the multi-output time series composed of the LAI and fAPAR data for Spain and Italy (i.e., the number of outputs is $Q=4$) from the beginning of 2003 until the end of 2013. We have experimented with a variable number of latent forces, $R \in \{1,2,3\}$, yet using more than $R=1$ LFs the model tends to overfit. Therefore, all the results displayed in the sequel correspond to a single LF (i.e., $R=1$). We use all the data available from the MODIS sensor ($N=506$ samples per time series in total) without removing any data, except for truly missing data (marked with negative values in the original time series).

The recovered latent force (LF) and two examples of the modelled time series are displayed in Fig. \ref{fig:res_noMissing}, where we can see the model has succeeded in capturing the dynamics of the data by using a single LF. A quantitative measure of performance is provided by the mean squared error (MSE),
\begin{equation}
	\mse_q = \frac{1}{N_q}\sum_{i=0}^{N_q-1}{(y_{q,i}-\hat{y}_{q,i})^2},
\label{eq:mse}
\end{equation}
and the normalised MSE,
\begin{equation}
	\rmse_q(\%) = \frac{\mse_q}{\frac{1}{N_q}\sum_{i=0}^{N_q-1}{y_{q,i}^2}} \times 100,
\label{eq:rmse}
\end{equation}
where $y_q[n]$ denotes the true value of the $n$-th sample from the $q$-th time series, $\hat{y}_q[n]$ is the value predicted by the model, $N_q \le N$ is the number of samples available for the $q$-th time series ($N_q < N$ if there are true missing data that have been removed) and $q=1,\ldots,Q$.

In this case, for Spain we have MSE = 0.1854 (NMSE = 3.38\%) and MSE = 0.0080 (NMSE = 4.00\%) for LAI and fAPAR respectively, whereas for Italy we have MSE = 0.2263 (5.58\%) and MSE = 0.0043 (NMSE = 2.34\%), again for LAI and fAPAR respectively, revealing good modelling results in all cases.
\begin{figure}[!tb]
	\centering
	\includegraphics[width=0.45\textwidth]{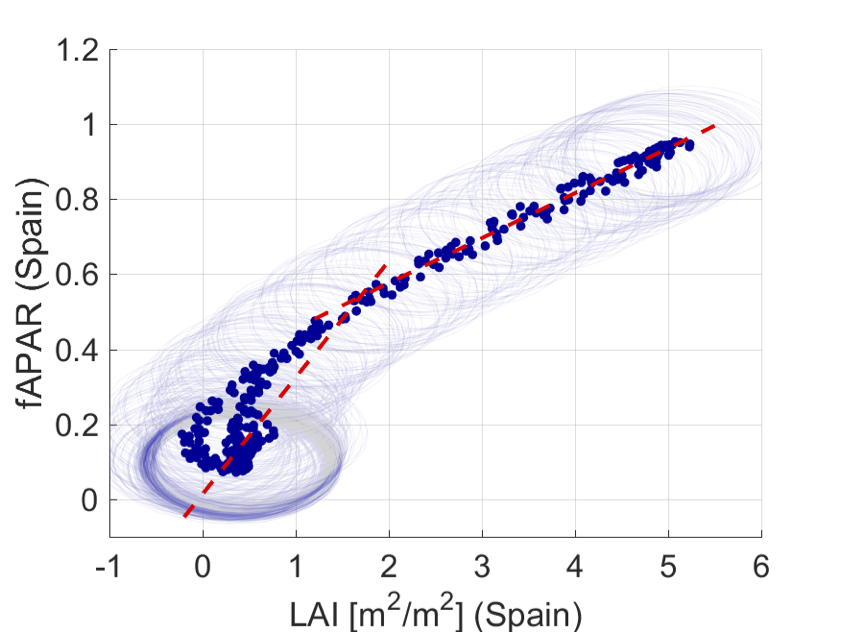}
\caption{fAPAR vs. LAI for the data learned using $R=1$ latent force and all the available data for years 2003--2013. We adjust two linear models for the first ($0-2$) and second ($2-5$) ranges of LAI values. The first model has an slope of $0.31~[m^2/m^2]^{-1}$, and the second $0.12~[m^2/m^2]^{-1}$.}
\label{fig:res_LaiFapar}
\end{figure}

\subsubsection{Learning LAI-fAPAR relations}

Note that the LAI-fAPAR relation is often assumed to be exponential, $\fapar = 1-\exp(-\alpha\cdot\lai/(\beta\cdot\cos(\text{SZA})))$, where $\alpha$ is associated with the composition of direct solar radiation and diffuse skylight as well as the multi-scattering within the canopy layer, and $\beta$ is related to the vegetation structure. We explore here the LAI-fAPAR relationship learned by the model. Figure \ref{fig:res_LaiFapar} displays the LAI vs. fAPAR scatter plot, obtained from the modelled time series for Spain.
%The shaded area corresponds
The ellipses around data samples show the uncertainty, that appears now in both axis as a consequence of the modelling uncertainty in both LAI and fAPAR.
Assuming an exponential  model of the form $\fapar = 1-\exp(\alpha\cdot\lai)$ we obtain a value of  $\alpha=-0.47$ by fitting a simple linear regression in the log-domain. This is a reasonable relation previously reported in the literature \cite{Ruimy1999}.
Interestingly, this nonlinear relation only holds in low vegetated instants, i.e. LAI$<$1, a piecewise linear slope of 0.31~[$m^2/m^2$] was observed (see fitted lines in Fig.~\ref{fig:res_LaiFapar}). For highly vegetated periods, LAI$>$2, the relation turns to be almost linear for the model with a lower slope of approximately 0.12~[$m^2/m^2$]. This simple observation could serve in the future to model the LAI-fAPAR relation combining the GP model with parametric linear activation functions, in a simple hybrid modeling scheme.

\begin{figure*}[!t]
	\centering
	\includegraphics[width=0.32\textwidth]{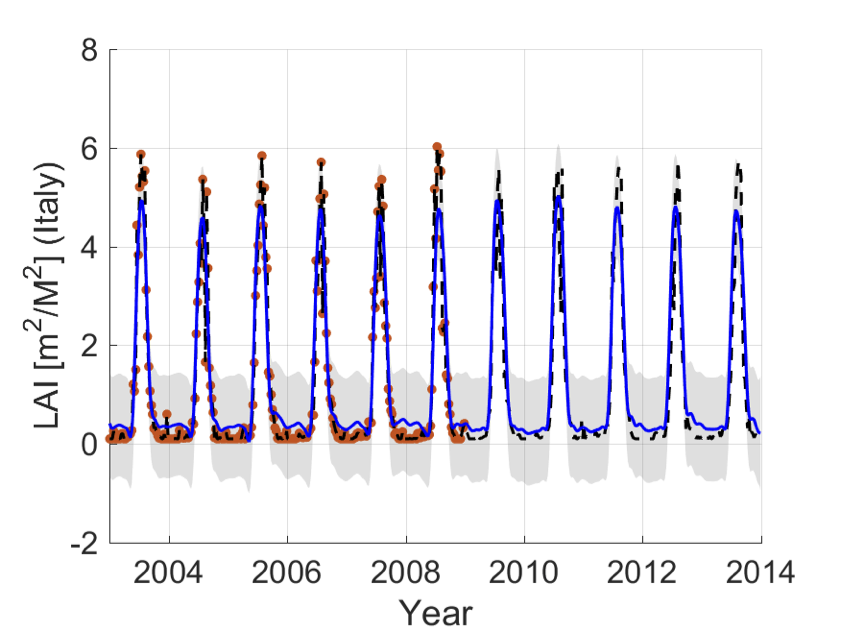}
	\includegraphics[width=0.32\textwidth]{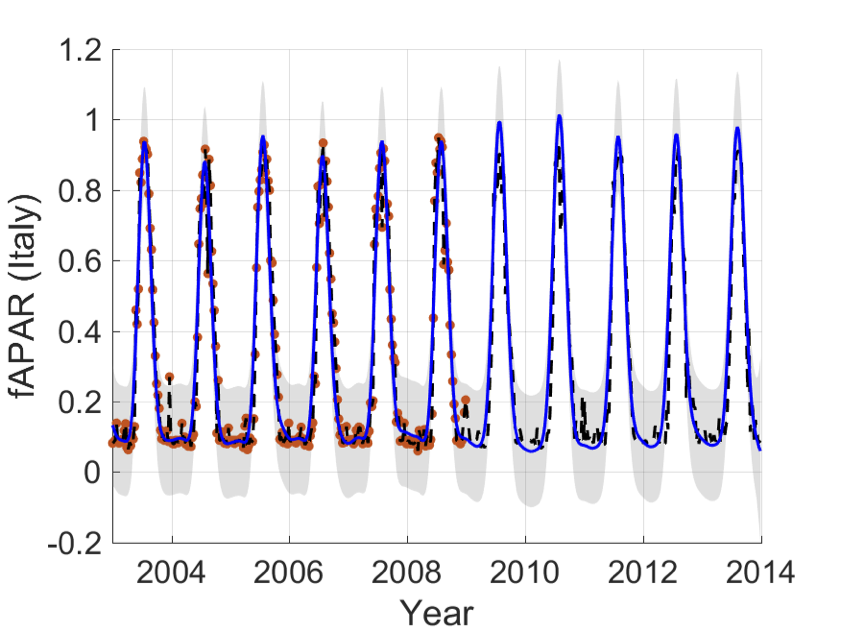}
	\includegraphics[width=0.32\textwidth]{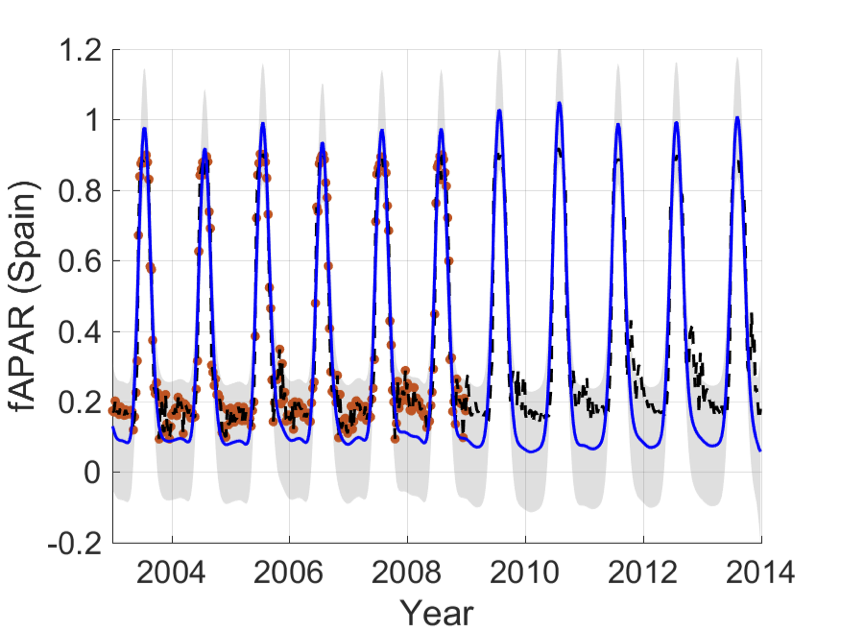}
\caption{Gap filling example using a single latent force (i.e., $R=1$) and all the LAI data from Spain (years 2003--2013), and removing the second half (years 2009--2013) of the other three time series: LAI (IT), fAPAR (IT) and fAPAR (ES). Training data (red dots), test data (black dashed line), predicted time series (blue line) and uncertainty measured by $\pm 2$ standard deviations about the mean predicted value (grey shaded area).}
\label{fig:res_missing}
\end{figure*}

\subsubsection{Multisite gap filling and domain adaptation}

Let us now investigate the capability of the method for recovering missing data (i.e., filling gaps in the measurements) by exploiting the information from other sites and times where such information is available. This can be cast as a problem of {\em domain adaptation} in space and time \cite{csurka2017domain}. 

{
We conducted two experiments. In both we simulated a scenario with different amounts of missing data. In the first one, we use all the available data from LAI (ES) and removed complete years from the other three time series (fAPAR (ES), LAI (IT) and fAPAR (IT)). We start by removing the data from year 2009 and end up removing the data from the years 2009--2013 (i.e., almost half of the available data from the last three time series). The modelled time series are displayed in Fig. \ref{fig:res_missing} for LF=1 (assuming one latent force only), whereas Table \ref{tab:mse_missing} shows the normalized MSE for all validation years and both sites using different methods and parameters: LFM with Gaussian-Gaussian kernels using one or two latent forces (LF), and LMC with $R=1$ using the Squared Exponential (SE) kernel and LMC using a composite sum kernel of the SE and a periodic kernel \cite{Rasmussen06,salcedo2014prediction}.
Figure \ref{fig:res_missing} shows that the LFM model is able to capture the correlations among all the time series, and reconstruct the second half of the last three outputs when only the first one is available for that time interval. Moreover, the error always remains within the same order of magnitude, even when we completely removed the second half of the data from the last three time series. This illustrates the robustness of the proposed approach w.r.t. large amounts of missing data, its ability to import information across sites and time, and its suitability for gap filling problems in Earth observation applications.
}

Table~\ref{tab:mse_missing} shows that %for this particular problem 
better results are obtained when using LFM with two LFs, although when many years are missing (last rows) results can be considerably worse (i.e. LAI (IT)), probably due to the complexity of the model and a certain risk of overfitting. We can also see that the LMC model generally gets better results in this experiment. This shows that having an overparameterized and flexible model such as LFM in this case may not be a good strategy to tackle problems with clearly dominating modes of variation, in this case the seasonal cycle.

\begin{table}[!htb]
	\small
	\setlength{\tabcolsep}{2pt}
	\centering
		\caption{ Normalized MSE for the missing data (in the test set) using LFM with one or two latent forces (LF), and LMC with $R=1$ using a Square Exponential kernel, or a composite of SE and periodic kernels (SE+Per).}
	\label{tab:mse_missing}
	\begin{tabular}{|c|c|c|c|}
		\hline
		Missing years & LAI (IT) & fAPAR (IT) & fAPAR (ES) \\
		\hline
		\multicolumn{4}{c}{LFM-GG, LF=1} \\ \hline
        2009       & 5.99 \% & 2.56 \% & 4.16 \% \\
        2009--2010 & 6.03 \% & 2.73 \% & 4.40 \% \\
        2009--2011 & 6.16 \% & 2.84 \% & 4.53 \% \\
        2009--2012 & 6.48 \% & 3.11 \% & 4.76 \% \\
        2009--2013 & 6.73 \% & 3.35 \% & 5.00 \% \\
		\hline
		\multicolumn{4}{c}{LFM-GG, LF=2} \\ \hline
		2009       & 4.20 \% & 2.75 \% & 0.76 \% \\
        2009--2010 & 4.19 \% & 2.79 \% & 1.03 \% \\
        2009--2011 & 4.45 \% & 2.83 \% & 1.47 \% \\
        2009--2012 & 6.55 \% & 2.36 \% & 1.37 \% \\
        2009--2013 & 8.81 \% & 6.50 \% & 12.54 \% \\
        \hline
		\multicolumn{4}{c}{LMC, $R=1$} \\ \hline
        2009       & 4.36 \% & 1.74 \% & 1.23 \% \\
        2009--2010 & 4.23 \% & 2.46 \% & 1.95 \% \\
        2009--2011 & 4.49 \% & 3.48 \% & 3.00 \% \\
        2009--2012 & 4.87 \% & 4.18 \% & 4.15 \% \\
        2009--2013 & 5.50 \% & 4.80 \% & 5.55 \% \\
        \hline
        
        % SE+RBF periodic
        \multicolumn{4}{c}{LFM: SE+Per + GG, LF=1} \\ \hline
        2009      & 5.99 \% & 2.56 \% & 4.15 \% \\
        2009-2010 & 6.02 \% & 2.73 \% & 4.40 \% \\
        2009-2011 & 6.17 \% & 2.85 \% & 4.54 \% \\
        2009-2012 & 6.48 \% & 3.11 \% & 4.76 \% \\
        2009-2013 & 6.73 \% & 3.35 \% & 4.99 \% \\
        \hline
        \multicolumn{4}{c}{LMC SE+Per, $R=1$} \\ \hline
        2009      & 4.35 \% & 1.74 \% & 1.23 \% \\
        2009-2010 & 4.22 \% & 2.46 \% & 1.95 \% \\
        2009-2011 & 4.49 \% & 3.49 \% & 3.00 \% \\
        2009-2012 & 4.87 \% & 4.19 \% & 4.15 \% \\
        2009-2013 & 5.50 \% & 4.79 \% & 5.55 \% \\

        \hline
	\end{tabular}
\end{table}

In the second experiment, we used LFM and LMC multi-output models working on either LAI or fAPAR separately instead. As in the previous experiment, we used the whole time series of Spain %untouched (not removing any years), 
while removing data from the years 2009-2013 in the Italy time series. Table~\ref{tab:mse_missing_sep_vars} shows the results on the predicted time series for Italy only (the one where training data were removed). Higher accuracy are obtained %Results are better 
when we use different models to handle LAI and fAPAR separately over using 
a single multi-output model to handle all variables (LAI and fAPAR) at the same time, as done in the previous experiment. This confirms that while LFM can cope with large multi-output problems (accuracy does not decrease dramatically), it is better to split the problem in subsets of related variables which helps modeling. Also compared with the previous experiment, when models deal with only one physical variable (LAI or fAPAR) the LMC model obtains better results than the LFM using one LF, and similar ones than LFM with two LFs. We want to emphasize here that we have a fitting-vs-interpretability choice to make in general: while an LMC might yield a slightly better predictive accuracy in some problems, LFM can yield useful insight on the problem such as the learned latent forces or hyperparameters with physical meaning.

\begin{table}[!htb] 
	\small
	\setlength{\tabcolsep}{2pt}
	\centering
		\caption{ Normalized MSE for the missing data (in the test set) on multi-output models trained on LAI and fAPAR separately using LFM and LMC with different $R$ values.} %with 1 or 2 LFs, and LMC with $R=1$.}
	\label{tab:mse_missing_sep_vars}
	\begin{tabular}{|c|c|c|}
		\hline
		Missing years & LAI (IT) & fAPAR (IT) \\
		\hline
        % LFM, $R=1$ & 1 LFs, LAI       & 1 LFs, fAPAR     \\ \hline
		LFM, LF=1 &    & \\ \hline
        2009      & 4.98 \% & 2.39 \% \\
        2009-2010 & 4.98 \% & 2.59 \% \\
        2009-2011 & 5.14 \% & 3.12 \% \\
        2009-2012 & 6.56 \% & 3.59 \% \\
        2009-2013 & 6.73 \% & 4.14 \% \\
        \hline
        %         & 2 LFs, LAI       & 2 LFs, fAPAR \\ \hline
	    LFM, LF=2  &    & \\ \hline
        2009       & 3.77 \% & 1.31 \% \\
        2009--2010 & 3.85 \% & 1.58 \% \\
        2009--2011 & 3.97 \% & 1.89 \% \\
        2009--2012 & 4.46 \% & 2.30 \% \\
        2009--2013 & 4.52 \% & 2.69 \% \\
        \hline
        LMC, $R=1$ &    & \\ \hline                  
	    % & LMC $R=1$, LAI   & LMC $R=1$, fAPAR \\ \hline
        2009      & 3.98 \% & 1.30 \% \\
        2009-2010 & 3.94 \% & 1.57 \% \\
        2009-2011 & 4.16 \% & 1.99 \% \\
        2009-2012 & 4.63 \% & 2.36 \% \\
        2009-2013 & 5.31 \% & 2.78 \% \\
        \hline        
	\end{tabular}
\end{table}

\subsubsection{Dealing with high missing data rates}

\begin{figure*}[!t]
	\centering
	\includegraphics[width=0.32\textwidth]{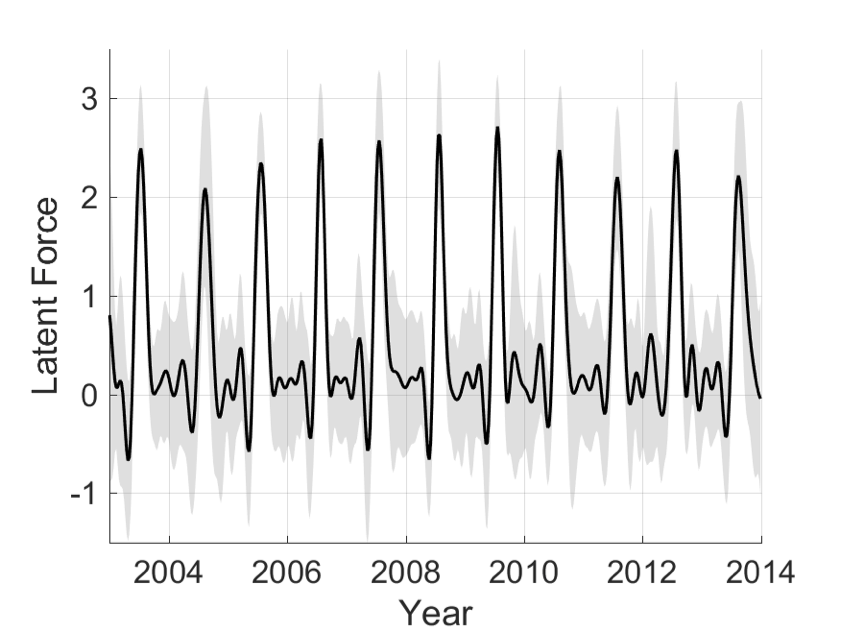}
	\includegraphics[width=0.32\textwidth]{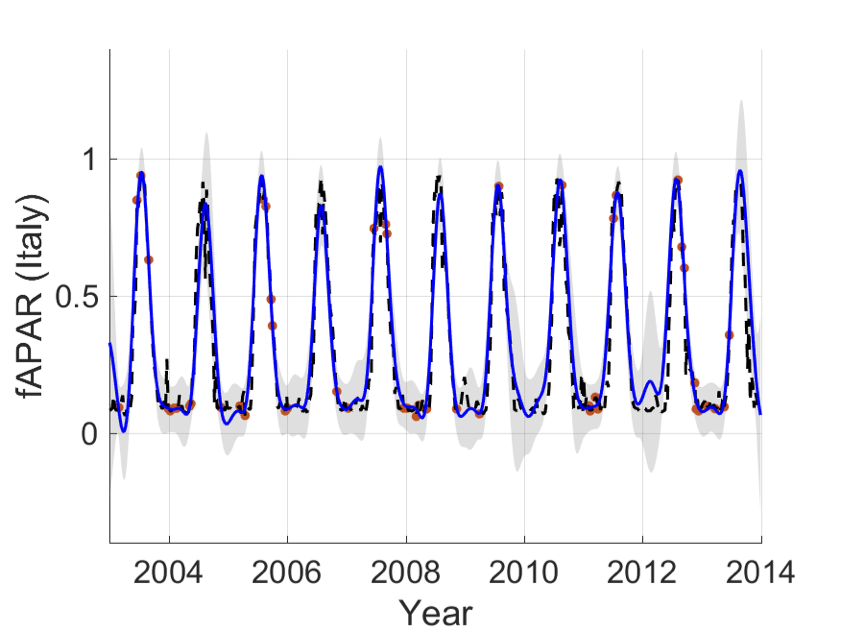}
	\includegraphics[width=0.32\textwidth]{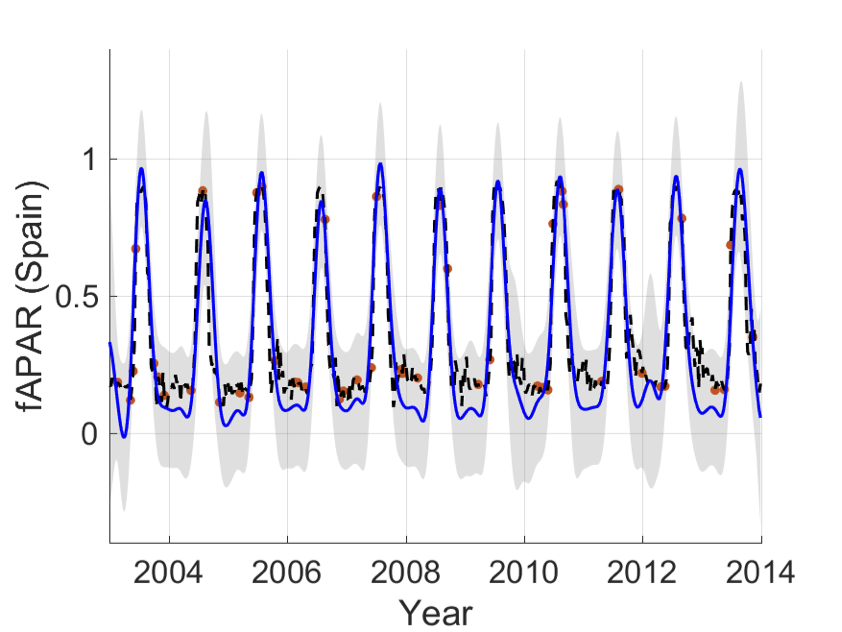}
\caption{Example of a high missing data rate scenario (90\% of the data are missing) using a single LF (i.e., $R=1$). Training data (red dots), test data (black dashed line), predicted time series (blue line) and uncertainty measured by $\pm 2$ standard deviations about the mean predicted value (grey shaded area).}
\label{fig:res_missing2}
\end{figure*}

Fig. \ref{fig:res_missing2} shows an example of the fAPAR obtained when 90\% data are missing. Even in this extreme case (where 456 and 453 out of the 506 available samples are missing for Italy and Spain, respectively) the method is able to capture the underlying dynamics of the multi-output time series. Notice also that, even in the large uncertainty regions, where no samples at all are available, the predicted value is not far from the true value.

\section{Conclusions}
\label{sec:conclusions}
We introduced process convolutions and latent force models in particular as a powerful machine learning framework for hybrid modeling. The framework exploits both observational data and a mechanistic model in the form of differential equations. The LFM is based on Gaussian processes, and enables us to perform gap filling of multiple time series importing information across time and space efficiently. Being based on solid Bayesian statistics, the GP modeling generally yields efficient and accurate parameter estimates, accompanied by confidence intervals for the estimations, and since the kernel is derived from a relevant ODE, the predictions are forced to respect a prior knowledge about the dynamic underlying process governing the system. In turn, the optimization leads to learning these latent forces automatically from data, as well as the hyperparameters of the model which account for physical quantities.

Our first study showcased the situation where a clear (yet limited) understanding is available, and considered SM time series acquired with different satellite microwave sensors. The LFM allowed us to seamlessly transfer the information across satellite records to reconstruct short and long data gaps. Furthermore, by incorporating a first order differential equation in the GP model, we showed how one can automatically learn ODE parameters related to soil moisture persistence and discover precipitation-related driving forces. 
In our second example we illustrated the performance in multiple sites gap filling LAI and fAPAR from MODIS time series exploiting spatio-temporal correlations. This has been shown to be an important problem for monitoring of vegetation from optical remote sensing data. In this case of data scarcity, uneven sampling and lack of mechanistic model of the LAI-fAPAR relations, one can assume a flexible LFM formed by Gaussian latent forces and coupling mechanism, which provide high resolution and accurate time series predictions. 

The proposed approach reconciles the two main approaches in remote sensing parameter estimation by blending statistical learning and mechanistic models. Our future work is related to tackling more challenging problems in geoscience and remote sensing: coupling of different (partially known) ODEs, study the error propagation and parameters distribution, and including non-linear differential equations and terms.

\bibliographystyle{IEEEtran}
\bibliography{ALL_urlfree,daniel,refs,references,acmart}

\end{document}